\documentclass[lettersize,journal]{IEEEtran}
\usepackage{amsmath,amsfonts,amssymb}
\usepackage{algorithmic}
\usepackage{algorithm}
\usepackage{array}
\usepackage{textcomp}
\usepackage{stfloats}
\usepackage{url}
\usepackage{verbatim}
\usepackage[pdftex]{graphicx}
\usepackage{comment}
\usepackage{cite}
\usepackage{booktabs}
\usepackage{makecell}
\usepackage{multirow}
\newtheorem{theorem}{Theorem}

\newtheorem{definition}[theorem]{Definition}

\usepackage{subfigure}

\usepackage{ifpdf}

\usepackage[dvipsnames]{xcolor}

\usepackage[normalem]{ulem}
\useunder{\uline}{\ul}{}

\usepackage{pifont}

\begin{document}

\title{3DeepRep: 3D Deep Low-rank Tensor Representation for Hyperspectral Image Inpainting}

\author{Yunshan Li, Wenwu Gong, Qianqian Wang, Chao Wang, Lili Yang
 ~\IEEEmembership{\\Department of Statistics and Data Science\\Southern University of Science and Technology\\Shenzhen, China\\ 
 E-mail: lyunshan565@gmail.com; gongww@sustech.edu.cn; 12331103@mail.sustech.edu.cn; wangc6@sustech.edu.cn; yangll@sustech.edu.cn}
\thanks{This research was supported by the Educational Commission of Guangdong Province (Grant No. 2021ZDZX1069), the SUSTech Presidential Postdoctoral Fellowship, and the Key Laboratory of TechFin in SUSTech. (Yunshan Li and Wenwu Gong are co-first authors.) (Corresponding authors: Chao Wang; Lili Yang.)}
}

\markboth{T\MakeLowercase{his work has been submitted to the IEEE for possible publication.} C\MakeLowercase{opyright may be transferred without notice, after which this version may no longer be accessible.}}%
{Shell \MakeLowercase{\textit{et al.}}: A Sample Article Using IEEEtran.cls for IEEE Journals}


\maketitle

\begin{abstract}
Recent approaches based on transform-based tensor nuclear norm (TNN) have demonstrated notable effectiveness in hyperspectral image (HSI) inpainting by leveraging low-rank structures in latent representations. Recent developments incorporate deep transforms to improve low-rank tensor representation; however, existing approaches typically restrict the transform to the spectral mode, neglecting low-rank properties along other tensor modes. In this paper, we propose a novel \underline{3}-directional \underline{deep} low-rank tensor \underline{rep}resentation (3DeepRep) model, which performs deep nonlinear transforms along all three modes of the HSI tensor. To enforce low-rankness, the model minimizes the nuclear norms of mode-$i$ frontal slices in the corresponding latent space for each direction ($i=1,2,3$), forming a 3-directional TNN regularization. The outputs from the three directional branches are subsequently fused via a learnable aggregation module to produce the final result. An efficient gradient-based optimization algorithm is developed to solve the model in a self-supervised manner. Extensive experiments on real-world HSI datasets demonstrate that the proposed method achieves superior inpainting performance compared to existing state-of-the-art techniques, both qualitatively and quantitatively.
\end{abstract}

\begin{IEEEkeywords}
Deep transform, 3-directional tensor nuclear norm (3DTNN), hyperspectral image inpainting, low-rank tensor representation, self-supervised learning.
\end{IEEEkeywords}
\section{Introduction} 
Hyperspectral images (HSIs), distinguished by their high spectral resolution encompassing numerous contiguous spectral bands, have become indispensable across various scientific and practical domains. For example, in urban planning, HSIs provide detailed spectral data crucial for land cover classification and monitoring of urban expansion \cite{Khan2018Modern}; in mineral exploration, these images facilitate mineral identification and mapping \cite{Booysen2021Accurate}; in precision agriculture, HSIs have revolutionized methodologies related to crop monitoring, soil evaluation, and pest management \cite{Ram2024A}. Despite their wide applicability, the practical utility of HSIs is frequently hindered by data degradation, including missing or corrupted spectral-spatial information. These degradations often arise due to the sensor malfunctions, atmospheric disturbances (e.g., clouds, haze), or transmission errors, which negatively impact downstream tasks such as classification, target detection, and change detection \cite{Sun2021TargetDT, Imani2020AnOO, Han2022MultimodalHU}. Consequently, hyperspectral image inpainting, aimed at reconstructing incomplete or corrupted HSI data, has become an essential preprocessing step \cite{Bhargava2024Hyperspectral}.

HSI inpainting is an ill-posed problem. A common strategy for addressing this issue is leveraging prior knowledge, such as low-rankness \cite{Hu2017TheTT, Liu2009TensorCF, Lu2018TensorRP}. An HSI is typically represented as a 3-mode tensor $\mathcal{X}\in \mathbb{R}^{n_1\times n_2 \times n_3}$, where the first two modes are spatial dimensions and the third mode is spectral \cite{Kolda2009TensorDA, Yaman2020LowRankTM, Li2018FusingHA}. The low-rank inpainting problem can be formulated as \cite{luo2022self}
\begin{align}
\min_{\mathcal{X}}\,\mathrm{rank}(\mathcal{X})+\lambda\mathbf{\mathit{L}}(\mathcal{X},\mathcal{O}),
\label{eq:LR}
\end{align}
where $\mathcal{X}$ is the recovered tensorial image, $\mathcal{O}$ is the observed tensorial image
$\mathbf{\mathit{L}}(\mathcal{X},\mathcal{O})$ is the data fidelity term, $\lambda$ is the hyperparameter to balance the data-fidelity term and the regularization term. However, the notion of tensor rank is not unique \cite{Kolda2009TensorDA, Dian2019HyperspectralIS}, leading to various tensor decompositions and corresponding rank definitions. For instance, the CANDECOMP/PARAFAC (CP)-rank \cite{Carroll1970AnalysisOI, Harshman1970FoundationsOT}, is defined as the smallest number of rank-one tensors needed to exactly reconstruct the original tensor. However, determining the CP-rank precisely is  NP-hard, making practical applications challenging. Tucker-rank \cite{Tucker1966SomeMN} is defined as a tuple representing the ranks of the mode-wise unfolding of the tensor, but may not capture cross-mode correlations effectively. Moreover, numerous ranks based on tensor network decomposition are proposed. Tensor Train (TT)-rank \cite{Oseledets2011TensorTrainD} offers efficiency but struggles with complex dependencies due to its linear structure. Tensor Ring (TR)-rank \cite{Zhao2016TensorRD} supports cyclic correlations, yet its fixed topology limits flexibility. Fully-connected Tensor Network (FCTN)-rank \cite{Zheng2021FullyConnectedTN, Zheng2022TensorCV} enhances representational capacity, albeit at the cost of high complexity.

Recently, tensor tubal-rank \cite{Liu2016LowTubalRankTC} has gained attention for HSI inpainting \cite{Xiao2020LowRankPT, Dian2019HyperspectralIS, Xu2019NonlocalPT}, which is based on the tensor singular value decomposition (t-SVD). A convex surrogate, the tensor nuclear
norm (TNN) \cite{Zhang2014NovelMF} is proposed to address the problem that minimizing tubal rank is NP-hard. Applying TNN as the measure of the tensor rank leads to the following model \cite{luo2022self}:
\begin{align}
\min_{\mathcal{X}}\;\sum_k\|(\mathcal{\widehat{X}})_3^{(k)}\|_*+\lambda\mathbf{\mathit{L}}(\mathcal{X},\mathcal{O}) \quad
\mathrm{s.t.\;} \ \mathcal{\widehat{X}} = f(\mathcal{X}),
\label{eq:TNN0}
\end{align}
where $(\cdot)_3^{(k)}$ represents the $k$-th mode-3 frontal slice \cite{zheng2019mixed}, $f(\cdot)$ is the discrete Fourier transform (DFT) along the third mode. The t-SVD framework interprets Fourier-transformed spectral correlations via SVDs on spatial slices, guiding the design of $g(\cdot)$ to produce latent representations. That is, the model \eqref{eq:TNN0} is equivalent to: 
\begin{align}
\min_{\mathcal{\widehat{X}}}\;\sum_k\|(\mathcal{\widehat{X}})_3^{(k)}\|_*+\lambda\mathbf{\mathit{L}}(\mathcal{X},\mathcal{O}) \quad
\mathrm{s.t.\;} \ \mathcal{X} = g(\mathcal{\widehat{X}}),
\label{eq:TNN}
\end{align}
where $g(\cdot)$ denotes the inverse discrete Fourier transform (IDFT) along the third mode, mapping the latent tensor $\mathcal{\widehat{X}}$ to the image domain \cite{Li2023H2TFFH}. The latent representation is constructed to be as low-rank as possible, such that the regularization $\min \sum_k\|(\mathcal{\widehat{X}})_3^{(k)}\|_*$ captures the underlying structure effectively. The transform $g(\cdot)$ enables the recovery of more complex and informative image content from the low-rank domain, thus enhancing both the expressiveness and fidelity of the reconstruction \cite{Lu2019LowRankTC, luo2022self}. 

Over time, the transform $g(\cdot)$ along the third mode in the TNN framework has evolved from the IDFT to the general inverse linear transform, and then to the nonlinear transform, and even to the deep neural network. Initially, the inverse discrete Fourier transform (IDFT) was used, which was later replaced by the inverse discrete cosine transform (IDCT) \cite{Xu2019AFA, Madathil2018DCTBW}, and subsequently extended to other unitary transforms \cite{Song2020RobustTC}. To further increase model flexibility, researchers introduced non-invertible transforms such as the framelet transform \cite{Jiang2019FrameletRO}, data-dependent linear transforms \cite{Kong2019TensorQN}, and learnable dictionary-based linear transforms \cite{Jiang2020DictionaryLW}. Notably, all these variants retain the property of linearity. A shift to nonlinear and deep transform designs marks the next stage of development. NTTNN \cite{Li2021NonlinearTI} introduced a nonlinear spectral mapping to substitute for the Fourier transform. S2NTNN \cite{luo2022self} adopted a fully connected (FC) network along the spectral mode to realize a deep transform $g(\cdot)$, while CoNoT \cite{wang2022conot} further integrated an FC layer along the spectral mode with a convolutional neural network (CNN) operating over the spatial dimensions.

However, the aforementioned methods under the TNN framework are only concerned with the correlation along the third mode, ignoring the use of $g(\cdot)$ to characterize the correlations along the other two modes. As a result, they fail to capture the full multilinear structure of HSI data, neglecting correlations and low-rankness along the other two modes (i.e., spatial-spectral and spectral-spatial). We refer to this limitation as \textit{direction inflexibility}, where the rank prior and transform design are restricted to a single tensor mode.

\subsection{Contribution}
To address the problem of \textit{direction inflexibility}, we proposed a \textbf{3}-directional \textbf{deep} low-rank tensor \textbf{rep}resentation (3DeepRep) model, which conducts $g(\cdot)$'s along three directions respectively and then merges the recovery results from the three directions into a final recovered HSI image.  

\begin{figure*}[!htbp]
    \centering
\includegraphics[width=0.8\textwidth]{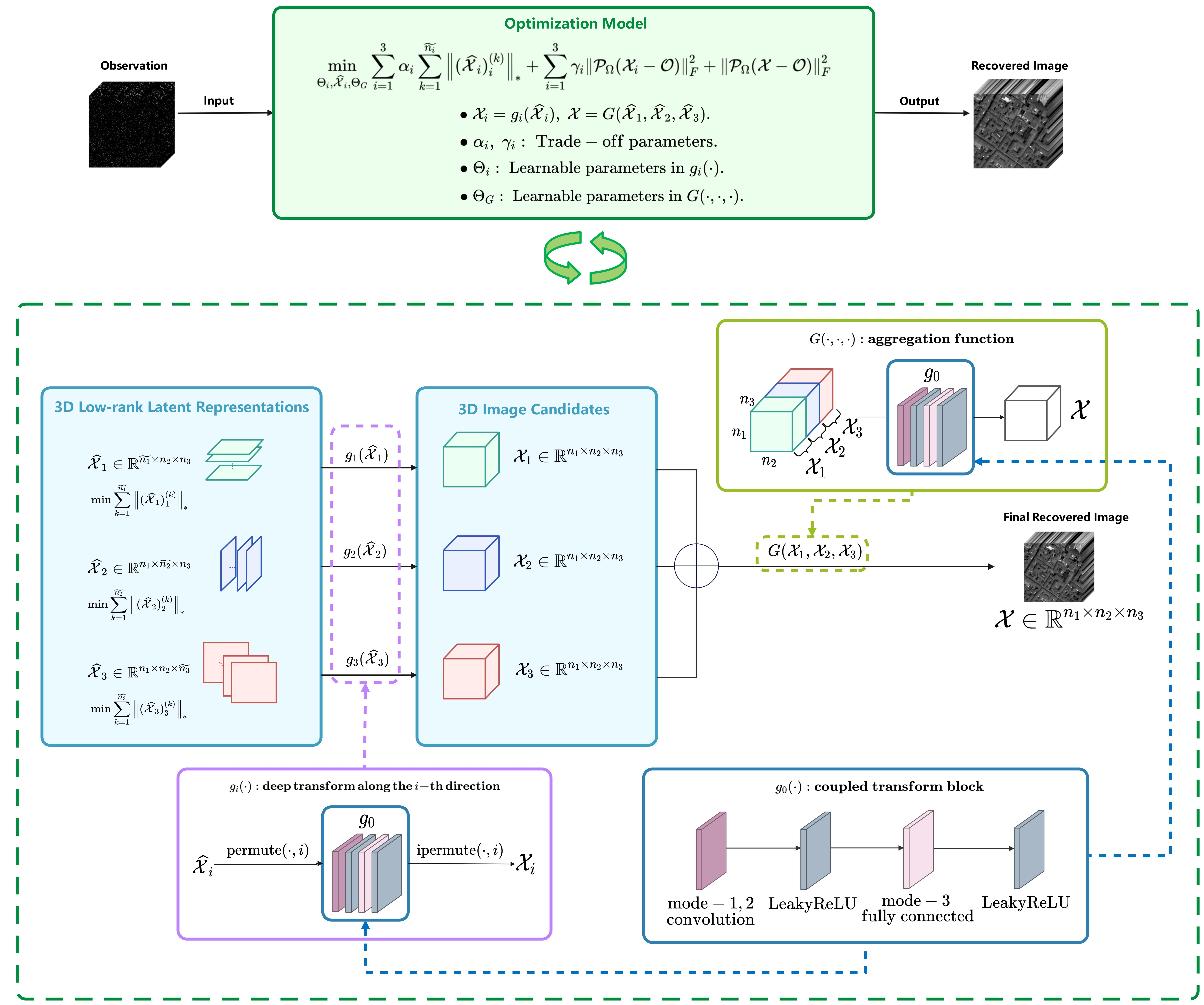}
    \caption{Flowchart of the proposed 3DeepRep method for HSI inpainting. The hyperspectral image is encoded into three low-rank latent tensors, where low-rankness is enforced via nuclear norm minimization on frontal slices along three modes. These tensors are then mapped to the original image domain through a \textit{coupled transform block} along the corresponding directions, producing three directional candidates. An \textit{aggregation module} fuses these into the final recovered image. This framework fully exploits spatial–spectral structures in all three tensor modes, addressing the directional limitations of previous methods.}
\label{fig:flowchart}
\end{figure*}

As illustrated in Figure~\ref{fig:flowchart}, we begin by assuming that the hyperspectral image exhibits low-rank structure in three respective latent spaces, denoted as $\mathcal{\widehat{X}}_1$, $\mathcal{\widehat{X}}_2$, and $\mathcal{\widehat{X}}_3$. For each latent tensor $\mathcal{\widehat{X}}_i$, low-rankness is enforced by minimizing the sum of nuclear norms across its frontal slices \cite{zheng2019mixed}. Each $\mathcal{\widehat{X}}_i$ is then mapped back to the original image domain through a deep transform $g_i(\cdot)$ designed along the $i$-th direction. The transform $g_i(\cdot)$ is implemented using a \textit{Coupled Transform Block} (CTB) \cite{wang2022conot}, which consists of a fully connected (FC) layer coupled with a 2D convolutional neural network (CNN). This structure ensures that $g_i(\cdot)$ captures both directional and cross-modal correlations effectively. The transformed outputs $\mathcal{X}_1 = g_1(\mathcal{\widehat{X}}_1)$, $\mathcal{X}_2 = g_2(\mathcal{\widehat{X}}_2)$, and $\mathcal{X}_3 = g_3(\mathcal{\widehat{X}}_3)$ serve as directional reconstructions of the original image. To fuse these directional reconstructions into a final high-quality image, we introduce an \textit{Aggregation Module} (AM) $G(\mathcal{X}_1, \mathcal{X}_2, \mathcal{X}_3)$, which is also built using a CTB. This module is designed to model the complex inter-directional dependencies among the three candidates and produce the final HSI inpainting result.

\begin{figure}[!htbp]
    \centering
    \includegraphics[width=0.45\textwidth]{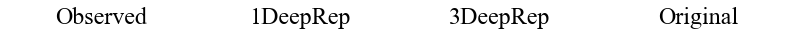}
    \includegraphics[width=0.45\textwidth]{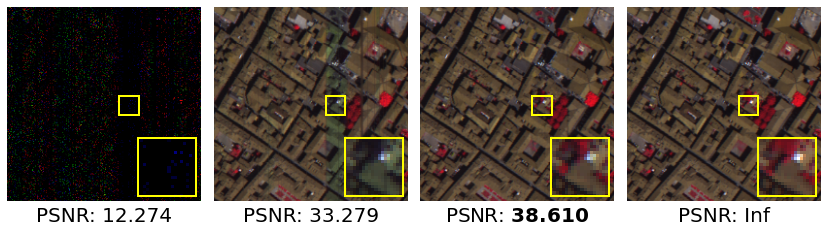}
    \includegraphics[width=0.45\textwidth]{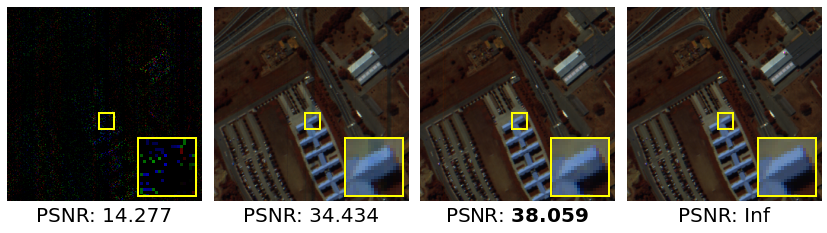}
    \includegraphics[width=0.45\textwidth]{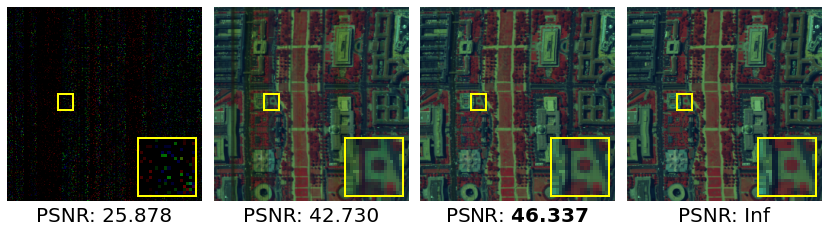}
    \caption{Qualitative and quantitative comparison of inpainting models: 1D-D-LRTR (CoNoT \cite{wang2022conot}) and proposed 3DeepRep. All the HSI images are false-color images composed of bands (R: 70, G: 40, and B: 10). From top to bottom are results on datasets of Pavia Centre, Pavia University, and Washington DC Mall, respectively. The data missing case is a mixed one with point missing, stripe missing and deadline missing (see Section \ref{sec:experiments}).}
\label{fig:intro}
\end{figure}

All latent variables $\mathcal{\widehat{X}}_i$ and the learnable parameters within $g_i(\cdot)$ and $G(\cdot, \cdot, \cdot)$ are jointly optimized in a self-supervised manner, without requiring access to ground-truth images. Fig. \ref{fig:intro} demonstrates the superiority of our proposed 3DeepRep method qualitatively and quantitatively compared to a 1DeepRep method, CoNoT \cite{wang2022conot}. The main contributions of this work are summarized as follows:

\begin{itemize}
    \item \textbf{Multi-directional transform-based low-rank tensor representation.} We extend the 1-directional coupled nonlinear transform-based low-rank measurement \cite{wang2022conot} to a 3-directional paradigm for HSI data. This multi-directional design enables a more comprehensive exploitation of the hyperspectral image’s low-rank structure and cross-modal correlations, thereby improving reconstruction accuracy. Furthermore, we treat the latent tensors $\mathcal{\widehat{X}}_i$'s as learnable variables directly. This design significantly reduces the number of learnable parameters compared with methods employing generators to produce latent tensors \cite{luo2022self,wang2022conot,NMTMA-17-727}, and simultaneously enhances the flexibility of the latent representations.
    
    \item \textbf{Coupled aggregation via deep transformation.} To fuse the 3-directional reconstructions, we introduce an aggregation function $G(\cdot, \cdot, \cdot)$ implemented by a coupled transform block. This block effectively captures the spatial-spectral interactions among the three candidates. The aggregation approach also significantly reduces the reliance on hand-crafted hyperparameters and improves computational efficiency, compared with traditional ADMM-based frameworks \cite{zheng2019mixed, NMTMA-17-727, Zheng2018TensorNR}.

    \item \textbf{Self-supervised optimization.} We develop an efficient gradient-based optimization algorithm tailored for our deep architecture. The entire framework, including the latent tensors, deep transforms, and aggregation module, is trained in a self-supervised manner without requiring ground truth. Experiments on real-world hyperspectral datasets demonstrate the superior performance of our method over existing state-of-the-art approaches.
\end{itemize}

\subsection{Organization}
The remainder of this article is organized as follows. Section \ref{sec:related work} introduces some related work. 
Section \ref{sec:pre} presents some preliminaries.
Section \ref{sec:method} proposes the 3DeepRep method step by step. Section \ref{sec:experiments} shows the experiment results and a discussion of the model. Section \ref{sec:conclusion} concludes this article. 

\section{Related Work}
\label{sec:related work}
\subsection{Deep Low-rank Latent Tensor-based Methods}
We use \textit{deep low-rank latent tensor-based methods} to refer to methods that consider the recovered tensorial image as the result of a transformation from a low-rank latent tensor, where the transformation is a deep neural network (named deep transform in this paper). This kind of method has attracted much attention these days since it not only leverages the low-rank property of images but also utilizes the strong expression ability of the deep neural network to preserve the complex information of images. Specifically, we summarize this kind of method as the following formula, which is a generalized form of (\ref{eq:LR}):
\begin{align}
\min_{\mathcal{X}}\,\mathbf{\mathit{L}}(\mathcal{X},\mathcal{O})\quad
\mathrm{s.t.\,}\mathcal{X} =  g_\theta(\mathcal{\widehat{X}}),\;\mathcal{\widehat{X}} \, \text{is low-rank},
\label{eq:DLRLT}
\end{align}
where $g_\theta(\cdot)$ denotes a deep neural network with learnable parameter $\theta$, which transforms the low-rank latent representation $\mathcal{\widehat{X}}$ into the original image space.

Some work uses the neural network to generate the low-rank latent $\mathcal{\widehat{X}}$.
In S2NTNN \cite{luo2022self}, the low-rank latent tensor $\mathcal{\widehat{X}}$ was generated by a two-layer mode-3 fully connected network and its low-rank property was enforced by minizing the nuclear norms $\sum_{k=1}^{\widetilde{n_3}}\|(\mathcal{\widehat{X}})_3^{(k)}\|_*$, where $(\cdot)_3^{(k)}$ means the $k$-th frontal slice, $\widetilde{n_3}$ is the number of frontal slices in $\mathcal{\widehat{X}}$. The deep transform $g_\theta(\cdot)$ had the same network structure of the generator of $\mathcal{\widehat{X}}$.
In CoNoT \cite{wang2022conot}, the low-rank latent tensor $\mathcal{\widehat{X}}$ was generated by a coupled nonlinear spatial-spectral transform and also was constrained by minimizing $\sum_{k=1}^{\widetilde{n_3}}\|(\mathcal{\widehat{X}})_3^{(k)}\|_*$. The deep transform $g_\theta(\cdot)$ had the same network structure of the generator of $\mathcal{\widehat{X}}$.
In \cite{Wang2025TensorCN}, the low-rank latent tensor $\mathcal{\widehat{X}}$ is constructed as a CANDECOMP/PARAFAC (CP) decomposition form to enforce its low-rankness with the factor matrices of the CP decomposition being the outputs of three multi-layer perceptrons (MLPs). The deep transform $g_\theta(\cdot)$ was designed as a U-net \cite{Ronneberger2015UNetCN}.

At the same time, some work does not design the low-rank latent tensor $\mathcal{\widehat{X}}$ as the output of a neural network generator. Instead, they endow $\mathcal{\widehat{X}}$ with a low-rank factorization/decomposition structure, directly setting the factorization/decomposition factors as learnable parameters. For example, Bacca et al. \cite{Bacca2021CompressiveSI} imposed a low-rank Tucker decomposition on $\mathcal{\widehat{X}}$, applying a multi-layer convolutional network as the deep transform $g_\theta(\cdot)$.
HLRTF \cite{Luo2022HLRTFHL} imposed a low-rank tensor factorization structure on $\mathcal{\widehat{X}}$, applying a two-layer mode-3 fully connected (FC) network as the deep transform $g_\theta(\cdot)$ as \cite{luo2022self}. And D-FCTN \cite{Su2025DeepFT} imposed a fully-connected tensor network decomposition structure on $\mathcal{\widehat{X}}$, applying a four-layer network as the deep transform $g_\theta(\cdot)$. These two works both learned the factors of the factorization/decomposition directly, but did not set them as the outputs of some generators. This kind of direct learning reduces the number of parameters introduced by generators and also increases the degree of freedom of the latent $\mathcal{\widehat{X}}$.

However, existing deep low-rank latent tensor methods typically apply the deep transform along a fixed mode. For example, S2NTNN \cite{luo2022self} and HLRTF \cite{Luo2022HLRTFHL} employ FC layers exclusively along the spectral mode. This fixed-mode strategy overlooks the potential benefits of extending deep transforms to other tensor modes, which can better capture cross-modal correlations and further exploit the low-rank properties of the latent tensor $\mathcal{\widehat{X}}$ from different perspectives. Consequently, such direction inflexibility may lead to suboptimal inpainting performance.

\subsection{3D Low-rank Tensor-based Methods}
We use \textit{3D low-rank tensor-based methods} to refer to methods that capture the low-rank property of the tensorial image from three different directions, respectively, by considering the recovered tensorial image as the result of three transformations from three low-rank latent representations, respectively. This kind of method characterizes different correlations on different modes by the transforms along different directions and fully leverages the low-rankness of images from different directions, which helps to recover images more effectively. Specifically, we summarize this kind of method as the following formula:
\begin{align*}
\min_{\mathcal{X}}\,\mathbf{\mathit{L}}(\mathcal{X},\mathcal{O}) \quad
\mathrm{s.t.}\, \mathcal{X} =  g_i(\mathcal{\widehat{X}}_i),\; \mathcal{\widehat{X}}_i \,\text{is low-rank}, \ i=1,2,3,
\end{align*}
where $g_i$ is a transform along the $i$-th mode/direction, which transforms the low-rank latent representation under the $i$-th direction, $\mathcal{\widehat{X}}_i$, into the original image space, $i=1,2,3$.
For instance, 3DTNN \cite{zheng2019mixed} and WSTNN \cite{Zheng2018TensorNR} employed the Fourier transform along the $i$-th mode as $g_i(\cdot)$, and enforced low-rankness by minimizing the nuclear norms of the mode-$i$ frontal slices of $\mathcal{\widehat{X}}_i$ for $i = 1, 2, 3$. Similarly, JRTQN \cite{Cheng2024TensorCV} adopted a data-dependent linear transform $Q_i^{\mathrm{T}}$ as $g_i(\cdot)$ and minimized the reweighted nuclear norms \cite{Huang2020LowRankAV} of mode-$i$ frontal slices of $\mathcal{\widehat{X}}_i$. 

However, all these methods rely solely on linear transforms for $g_i(\cdot)$, thereby failing to capture the complex nonlinear patterns inherent in image data. As a result, the recovered outputs are often less accurate and computationally efficient than those that could be obtained using deep neural networks.

\section{Preliminaries}
\label{sec:pre}
The primary notations used in this paper are introduced in Table \ref{tab: notation}. In addition, we introduce the tensor mode-$i$ permutation operator as follows.
\begin{table}[H]
\begin{center}
    \caption{Notations}
    \label{tab: notation}
    \begin{tabular}{cc}
    \toprule[1.5pt]
        \textbf{Notation} & \textbf{Explanation} \\ \hline
         $x,\mathbf{x},\mathbf{X},\mathcal{X}$& Scalar, vector, matrix, tensor.\\
         $\mathcal{X}(l,j,s)$& The ($l$, $j$, $s$)-th element of $\mathcal{X}$.\\
         $\mathcal{\widehat{X}}_i$ & Low-rank latent tensor under the $i$-th direction,  $i\in\{1,2,3\}$.\\
         $\mathcal{X}_i$ & Recovered HSI image from the $i$-th direction.\\
         $(\mathcal{\widehat{X}}_1)_1^{(k)}$& Mode-1 frontal slice\cite{zheng2019mixed} of $\mathcal{\widehat{X}}_1$: $\mathcal{\widehat{X}}_1(k,:,:)$. \\
         $(\mathcal{\widehat{X}}_2)_2^{(k)}$& Mode-2 frontal slice\cite{zheng2019mixed} of $\mathcal{\widehat{X}}_2$: $\mathcal{\widehat{X}}_2(:,k,:)$.\\
         $(\mathcal{\widehat{X}}_3)_3^{(k)}$& Mode-3 frontal slice\cite{zheng2019mixed} of $\mathcal{\widehat{X}}_3$: $\mathcal{\widehat{X}}_3(:,:,k)$. \\
         $\left \| \mathbf{X } \right \| _*$& The matrix nuclear norm of $\mathbf{X}$.\\
         $\left \| \mathcal{X } \right \| _F$& Frobenius norm: $\sqrt{\sum_{l,j,s}\mathcal{X }_{l,j,s}^2}$.\\
         $\times_i$& Mode-$i$ tensor-matrix product. \\
         $\circledast $ & 2D convolution operation.\\
         $w_1(\cdot)$ & A convolutional neural network (CNN) layer.\\
         $w_2(\cdot)$ & A fully-connected (FC) layer.\\
         $g_i(\cdot)$ & Deep transform along the $i$-th direction.\\
         $G(\cdot,\cdot,\cdot)$& Aggregation function.\\

    \bottomrule[1.5pt]
    \end{tabular}
\end{center}
\end{table}

\begin{definition} [Tensor Mode-$i$ Permutation\cite{zheng2019mixed}]
Let $\mathcal{X}\in \mathbb{R}^{m_1 \times m_2 \times m_3}$ be a third-order tensor. For any $i \in \{1,2,3\}$, we define its mode-$i$ permutation, denoted as $\operatorname{permute} (\mathcal{X},i)$, as a reordering of modes such that the $i$-th mode is moved to the third dimension, while the other two modes retain their original order in the first and second dimensions. Mathematically, the relationship between $\mathcal{X}$,  $\operatorname{permute} (\mathcal{X},1)$,  $\operatorname{permute} (\mathcal{X},2)$,  $\operatorname{permute} (\mathcal{X},3)$ is expressed as
\begin{align*}
\mathcal{X}(l,j,s) & \ = \ \operatorname{permute}(\mathcal{X},1) (j,s,l)\\
      & \ = \ \operatorname{permute}(\mathcal{X},2) (l,s,j)\\
      & \ = \ \operatorname{permute}(\mathcal{X},3) (l,j,s), 
        \\
        l\in [m_1],\,&j\in[m_2],\,s\in[m_3].
\end{align*}
The inverse operation of $\operatorname{permute}(\cdot,i)$ is denoted as $\operatorname{ipermute}(\cdot,i)$, $i\in\{1,2,3\}$.
\label{def:permute}
\end{definition}
\begin{definition}[Concatenation Function]
The concatenation of three tensors $\mathcal{X}_1, \mathcal{X}_2, \mathcal{X}_3$, each of dimension $n_1 \times n_2 \times n_3$, along the third mode is defined as follows:
\begin{align*}
\mathrm{concat}(\mathcal{X}_1, \mathcal{X }_2,\mathcal{X}_3)(l,j,s) = \begin{cases}
  \mathcal{X}_1(l,j,s_1), \ & s= s_1\\
  \mathcal{X}_2(l,j,s_2), \ & s= s_2+n_3^1\\
  \mathcal{X}_3(l,j,s_3), \ & s= s_3+n_3^1+n_3^2,
\end{cases}
\end{align*}
where $\mathrm{concat}(\mathcal{X}_1, \mathcal{X }_2,\mathcal{X}_3)\in\mathbb{R}^{n_1\times n_2\times 3n_3}$, $s_i \in [n_3]$, for $i=1,2,3$, $l\in [n_1]$, and $j\in [n_2]$.
\label{def:concat}
\end{definition}

\section{Proposed Method}
\label{sec:method}

\subsection{Coupled Transform Block}
We adopt the following module, referred to as the coupled transform block \cite{wang2022conot}, as the fundamental building unit for constructing deep transform layers:
\begin{equation}
    g_0(\cdot) = w_2\left(w_1 (\cdot)\right) ,
    \label{eq:g}
\end{equation}
where $w_1 (\cdot)$ denotes a slice-wise convolutional layer that captures spatial correlations within each frontal slice of the input tensor and $w_2 (\cdot)$ denotes a fully connected layer that models global dependencies along the spectral mode.

Specifically, let \( \widehat{\mathcal{X}} \in \mathbb{R}^{m_1 \times m_2 \times \widetilde{m_3}} \) denote the input latent tensor. The operator \( w_1(\cdot) \) applies a 2D convolution with a kernel size of \(3 \times 3\) to each frontal slice \(\widehat{\mathcal{X}}^{(k)}\), followed by an element-wise nonlinear activation. This operation can be expressed as
\begin{equation*}
    w_1(\widehat{\mathcal{X}}) = \sigma\left(\left\{ \mathbf{W}_{1}^{k} \circledast \widehat{\mathcal{X}}^{(k)} \right\}_{k=1}^{\widetilde{m_3}} \right),
\end{equation*}
where \(\mathbf{W}_1^k \in \mathbb{R}^{3 \times 3}\) is the convolution kernel for the \(k\)-th slice, and \(\circledast\) denotes the 2D convolution operator with zero-padding of 1 and stride of 1. The activation function \(\sigma\) is applied element-wise; in our implementation, we adopt LeakyReLU~\cite{he2015delving} as the activation function.

The subsequent FC layer \( w_2(\cdot) \) models inter-slice correlations by performing a mode-3 tensor-matrix product with a learnable weight matrix \(\mathbf{W}_2 \in \mathbb{R}^{m_3 \times \widetilde{m_3}}\). Formally, for the output tensor \( \widehat{\mathcal{Y}} = w_1(\widehat{\mathcal{X}}) \), we define:
\begin{equation*}
    w_2(\widehat{\mathcal{Y}}) = \sigma\left( \widehat{\mathcal{Y}} \times_3 \mathbf{W}_2 \right),
\end{equation*}
where \(\times_3\) denotes the mode-3 product between a tensor and a matrix, and \(\sigma\) is applied element-wise.

\subsection{The Structure of 3DeepRep}
\label{sec:1D_3D}

Motivated by enhancing the transform mechanism in the TNN framework, we propose a novel \textbf{3}-directional \textbf{deep} low-rank tensor \textbf{rep}resentation (3DeepRep) model. The key ideas are twofold: to construct directional deep transforms $g_i(\cdot)$ along each mode-$i$ ($i = 1,2,3$) based on a shared foundational transformation block $g_0(\cdot)$; integrate complementary information from the three directional transformations. We construct the directional deep transform $g_i(\cdot)$ along the $i$-th mode ($i=1,2,3$) as
\begin{equation}
g_i(\widehat{\mathcal{X}}_i) = \operatorname{ipermute}\left( g_0\left(\operatorname{permute}(\widehat{\mathcal{X}}_i, i) \right), i \right),
    \ i = 1,2,3,
    \label{eq:gi}
\end{equation}
where $\operatorname{permute}(\cdot)$ and $\operatorname{ipermute}(\cdot)$ are operations defined in Definition~\ref{def:permute}. Specifically, $g_i(\widehat{\mathcal{X}}_i)$ applies a CNN layer $w_1(\cdot)$ to the mode-$i$ frontal slices (i.e., slices over the two modes orthogonal to mode-$i$) to capture local correlations, followed by a FC layer $w_2(\cdot)$ applied along mode-$i$ to model the global correlation.

To integrate complementary information from the three directional transformations $g_1(\mathcal{\widehat{X}}_1), g_2(\mathcal{\widehat{X}}_2), g_3(\mathcal{\widehat{X}}_3)$, we define an aggregation function $G(\cdot,\cdot,\cdot) = g_3\left(\mathrm{concat}(\cdot,\cdot,\cdot)\right)$ that fuses the directional information to produce the final inpainting result: 
\begin{align}
    \mathcal{X}& \ = \ G(g_1(\mathcal{\widehat{X}}_1), g_2(\mathcal{\widehat{X}}_2), g_3(\mathcal{\widehat{X}}_3))\nonumber\\
    & \ = \ g_3\left( \mathrm{concat}(g_1(\mathcal{\widehat{X}}_1), g_2(\mathcal{\widehat{X}}_2), g_3(\mathcal{\widehat{X}}_3)) \right), \label{eq:G}
\end{align}
where $\mathrm{concat}(\cdot,\cdot,\cdot)$ is defined in Definition \ref{def:concat}.
The aggregation function treats the concatenated tensor as a hyperspectral image with $3n_3$ bands and applies the deep transform along the spectral dimension. This spectral-mode transformation is particularly effective for capturing inter-band dependencies across the three directional reconstructions, while adjusting the spectral dimension to match the target observation $\mathcal{O}$.

\subsection{3DeepRep for HSI Inpainting}
\label{sec:3D_model}

Based on the t-SVD framework defined in Eq.~\eqref{eq:TNN}, we formulate the directional optimization model for HSI inpainting along the $i$-th mode as:
\begin{small}
 \begin{equation}
    \min_{\Theta_i,\, \widehat{\mathcal{X}}_i}\, 
    \sum_{k=1}^{\widetilde{n}_i} \left\| (\widehat{\mathcal{X}}_i)_i^{(k)} \right\|_* + 
     \lambda_i \left\| \mathcal{P}_{\Omega}\left( g_i(\widehat{\mathcal{X}}_i) - \mathcal{O} \right) \right\|_F^2,
    \quad i = 1,2,3,
    \label{eq:modei-model}
\end{equation}   
\end{small}
where $\widehat{\mathcal{X}}_i$ denotes the low-rank latent tensor under the $i$-th direction, which is a 3-mode tensor, whose mode-$i$ frontal slices are assumed to be low-rank; $\widetilde{n}_i$ is the size of its $i$-th mode; $g_i(\widehat{\mathcal{X}}_i)$ is the reconstructed image from direction-$i$, hereafter denoted as $\mathcal{X}_i$; $\mathcal{O} \in \mathbb{R}^{n_1 \times n_2 \times n_3}$ is the partially observed HSI; $\Omega$ is the set of known index locations; $\mathcal{P}_{\Omega}(\cdot)$ denotes a projection operator that retains entries in $\Omega$ and zeros elsewhere; $\Theta_i$ represents the set of learnable parameters in $g_i(\cdot)$; and $\lambda_i$ balances the low-rank regularization and the data-fidelity term.

Each directional reconstruction $\mathcal{X}_i$ is treated as an image candidate contributing to the final estimation. Notably, the parameters $\Theta_i$ are learned in a self-supervised fashion, without requiring ground-truth supervision, by encouraging the latent representation $\widehat{\mathcal{X}}_i$ to be as low-rank as possible via the nuclear norm regularization term, inspired by \cite{luo2022self, wang2022conot}.

To build the complete recovery model, for any $i = 1,2,3$, we first define the directional sub-loss functions:
\begin{small}
\begin{align}
    \mathcal{L}_i(\Theta_i,\, \widehat{\mathcal{X}}_i) = 
    \sum_{k=1}^{\widetilde{n}_i} \left\| (\widehat{\mathcal{X}}_i)_i^{(k)} \right\|_* + 
    \lambda_i \left\| \mathcal{P}_{\Omega}\left( g_i(\widehat{\mathcal{X}}_i) - \mathcal{O} \right) \right\|_F^2.
    \label{eq:li}
\end{align}
\end{small}
Then, the proposed 3DeepRep model for HSI inpainting is formulated as:
\begin{small}
\begin{equation}
    \min_{\Theta_i,\, \widehat{\mathcal{X}}_i,\, \Theta_G} \;
    \sum_{i=1}^{3} \alpha_i\, \mathcal{L}_i(\Theta_i, \widehat{\mathcal{X}}_i) +
    \left\| \mathcal{P}_{\Omega} \left( G(\mathcal{X}_1, \mathcal{X}_2, \mathcal{X}_3) - \mathcal{O} \right) \right\|_F^2,
    \label{eq:3D-model}
\end{equation}
\end{small}
where $\{\alpha_i\}_{i=1}^3$ are weighting coefficients for each directional sub-loss, $G(\cdot,\cdot,\cdot)$ is the aggregation function defined in Eq.~\eqref{eq:G} that outputs the final recovered image, and $\Theta_G$ denotes the learnable parameters in $G(\cdot,\cdot,\cdot)$.

\textit{Remark:} In previous 3DTNN-based work \cite{zheng2019mixed,NMTMA-17-727,Zheng2018TensorNR}, it is assumed directly that $\mathcal{X}_1 = \mathcal{X}_2 = \mathcal{X}_3 = \mathcal{X}$, and the final $\mathcal{X}$ is obtained through the Alternating Direction Method of Multipliers (ADMM). However, this approach introduces three hyperparameters to balance the contributions among $\mathcal{X}_1$, $\mathcal{X}_2$, and $\mathcal{X}_3$, to choosing whom significantly increases the computational cost. In contrast, in this work, we employ an aggregation function $G(\cdot,\cdot,\cdot)$ to obtain the final $\mathcal{X}$, which eliminates the need to introduce additional hyperparameters, thereby reducing both the computational burden and the risk of errors caused by improper hyperparameter tuning.

\subsection{Gradient Descent-based Algorithm}
The proposed 3DeepRep model in (\ref{eq:3D-model}) exhibits strong nonlinear characteristics, making traditional optimization algorithms, such as the Alternating Direction Method of Multipliers (ADMM), ineffective for solving our model. Therefore, we employ gradient descent algorithms specifically designed for deep learning to update the learnable parameters $\{\Theta_i\}_{i=1}^3, \{\widehat{\mathcal{X}}_i\}_{i=1}^3, \Theta_G$. 
In this subsection, we derive the subgradient of the optimization loss in (\ref{eq:3D-model}), which contains special terms of the matrix nuclear norm $\|\cdot\|_*$, to show the validity of the gradient descent-based algorithm.

We rewrite the objective function in Equation (\ref{eq:3D-model}) as 
\begin{equation*}
\mathcal{L}=\sum_{i=1}^3\alpha_i\mathcal{L}_i + \mathcal{L}_4,
\end{equation*}
where $\{\mathcal{L}_i\}_{i=1}^3$ is defined in (\ref{eq:li}), and $\mathcal{L}_4 = \left\|\mathcal{P}_{\Omega}(G(\mathcal{X}_1, \mathcal{X}_2, \mathcal{X}_3)-\mathcal{O})\right\|_F^2$. We analyze the subgradients of $\mathcal{L}$ with respect to the learnable parameters $\{\Theta_i\}_{i=1}^3$, $\{\widehat{\mathcal{X}}_i\}_{i=1}^3$ and $\Theta_G$ respectively.

First, the subgradient of $\mathcal{L}$ with respect to $\Theta_i$, $i = 1,2,3$, is given by
\begin{align}
 \frac{\partial \mathcal{L}}{\partial \Theta_i} 
 &=\alpha_i \frac{\partial \mathcal{L}_i}{\partial \Theta_i}+\frac{\partial \mathcal{L}_4}{\partial \Theta_i} \nonumber\\ 
 &=2 \alpha_i \lambda_i \sum_{l, j, s}\left[\mathcal{P}_\Omega(\mathcal{X}_i-\mathcal{O})\right]_{l, j, s} \frac{\partial(\mathcal{X}_i)_{l, j, s}}{\partial \Theta_i} \nonumber\\ 
 &\quad +2 \sum_{l, j, s}\left[\mathcal{P}_\Omega(\mathcal{X}-\mathcal{O})\right]_{l, j, s} \frac{\partial \mathcal{X}_{l, j, s}}{\partial \Theta_i}.
\label{eq:wrt_thetai}
\end{align}
Next, we derive the subgradient of $\mathcal{L}$ with respect to $\widehat{\mathcal{X}}_i$, $i=1,2,3$. The following theorem is needed:
\begin{theorem}[Subgradient of the Nuclear Norm of a Matrix {\cite{Watson1992CharacterizationOT}}]
For a given matrix $\mathbf{X}$ with singular value decomposition expressed as $\mathbf{X} = \mathbf{U S V}^\top$, the subgradient of its nuclear norm can be characterized by:
\begin{align*}
 \frac{\partial\|\mathbf{X}\|_*}{\partial \mathbf{X}}=\left\{\widetilde{\mathbf{U}} \widetilde{\mathbf{V}}^{\top}+\mathbf{W} \mid \mathbf{U}^{\top} \mathbf{W}=0, \mathbf{W}\mathbf{V}=0,\|\mathbf{W}\|_{1} \leq 1\right\},   
\end{align*}
where $\widetilde{\mathbf{U}}$ and $\widetilde{\mathbf{V}}$ represent the matrices formed by the first $r$ singular vectors of $\mathbf{U}$ and $\mathbf{V}$, respectively, and $r = \mathrm{rank}(\mathbf{X})$.
\end{theorem}

In our algorithm, we apply the subgradient by setting $\mathbf{W} = \mathbf{0}$:
\begin{equation}
\widetilde{\mathbf{U}} \widetilde{\mathbf{V}}^{\top} \in \frac{\partial\|\mathbf{X}\|_*}{\partial \mathbf{X}}.
\label{eq:subg_nuc}
\end{equation}

We compute $\frac{\partial\sum_{k=1}^{\widetilde{n}_i} \left \| (\widehat{\mathcal{X}}_i )_i^{(k)} \right \|_* }{\partial (\widehat{\mathcal{X}}_i)_{l,j,s}}$ for each mode:

For $i=1$:
\begin{align}
\frac{\partial\sum_{k=1}^{\widetilde{n}_1} \left \| (\widehat{\mathcal{X}}_1 )_1^{(k)} \right \|_* }
{\partial (\widehat{\mathcal{X}}_1)_{l,j,s}} 
= (\widetilde{\mathbf{U}}^1_l \widetilde{\mathbf{V}}^1_l{}^{\top})_{j,s}, \label{eq:i1}
\end{align}

For $i=2$:
\begin{align}
\frac{\partial\sum_{k=1}^{\widetilde{n}_2} \left \| (\widehat{\mathcal{X}}_2 )_2^{(k)} \right \|_* }
{\partial (\widehat{\mathcal{X}}_2)_{l,j,s}} 
= (\widetilde{\mathbf{U}}^2_j \widetilde{\mathbf{V}}^2_j{}^{\top})_{l,s}, \label{eq:i2}
\end{align}

For $i=3$:
\begin{align}
\frac{\partial\sum_{k=1}^{\widetilde{n}_3} \left \| (\widehat{\mathcal{X}}_3 )_3^{(k)} \right \|_* }
{\partial (\widehat{\mathcal{X}}_3)_{l,j,s}} 
= (\widetilde{\mathbf{U}}^3_s \widetilde{\mathbf{V}}^3_s{}^{\top})_{l,j}, \label{eq:i3}
\end{align}

We also have:
\begin{small}
\begin{align}
\frac{\partial \left\|\mathcal{P}_\Omega(\mathcal{X}_i - \mathcal{O})\right\|_F^2}{\partial (\widehat{\mathcal{X}}_i)_{l,j,s}} 
= 2 \sum_{l^*, j^*, s^*} \left[\mathcal{P}_\Omega(\mathcal{X}_i - \mathcal{O})\right]_{l^*, j^*, s^*} \frac{\partial (\mathcal{X}_i)_{l^*, j^*, s^*}}{\partial (\widehat{\mathcal{X}}_i)_{l,j,s}}. \label{eq:f2}
\end{align}

\begin{align}
\frac{\partial \mathcal{L}_4 }{\partial (\widehat{\mathcal{X}}_i)_{l,j,s}} 
= 2 \sum_{l^*, j^*, s^*} \left[\mathcal{P}_\Omega(\mathcal{X} - \mathcal{O})\right]_{l^*, j^*, s^*} \frac{\partial \mathcal{X}_{l^*, j^*, s^*}}{\partial (\widehat{\mathcal{X}}_i)_{l,j,s}}. \label{eq:l4}
\end{align}
\end{small}
From (\ref{eq:i1})--(\ref{eq:l4}), we derive:
\begin{align}
\frac{\partial \mathcal{L}}{\partial (\widehat{\mathcal{X}}_i)_{l,j,s}} 
= \alpha_i d_i^1 + 2\alpha_i \lambda_i d_i^2 + 2 d_i^3, \label{eq:wrt_xi}
\end{align}
where
\begin{align*}
d_i^1 &= \begin{cases}
(\widetilde{\mathbf{U}}^1_l \widetilde{\mathbf{V}}^1_l{}^{\top})_{j,s}, & i=1, \\
(\widetilde{\mathbf{U}}^2_j \widetilde{\mathbf{V}}^2_j{}^{\top})_{l,s}, & i=2, \\
(\widetilde{\mathbf{U}}^3_s \widetilde{\mathbf{V}}^3_s{}^{\top})_{l,j}, & i=3;
\end{cases} \\
d_i^2 &= \sum_{l^*, j^*, s^*} \left[\mathcal{P}_\Omega(\mathcal{X}_i - \mathcal{O})\right]_{l^*, j^*, s^*} \frac{\partial (\mathcal{X}_i)_{l^*, j^*, s^*}}{\partial (\widehat{\mathcal{X}}_i)_{l,j,s}}; \\
d_i^3 &= \sum_{l^*, j^*, s^*} \left[\mathcal{P}_\Omega(\mathcal{X} - \mathcal{O})\right]_{l^*, j^*, s^*} \frac{\partial \mathcal{X}_{l^*, j^*, s^*}}{\partial (\widehat{\mathcal{X}}_i)_{l,j,s}}.
\end{align*}

Finally, the gradient with respect to $\Theta_G$ is given by:
\begin{align}
\frac{\partial \mathcal{L}}{\partial \Theta_G} 
&= \frac{\partial \mathcal{L}_4}{\partial \Theta_G} \nonumber\\
&= 2 \sum_{l, j, s} \left[\mathcal{P}_\Omega(\mathcal{X} - \mathcal{O})\right]_{l,j,s} \frac{\partial \mathcal{X}_{l,j,s}}{\partial \Theta_G}.
\label{eq:thetag}
\end{align}

With gradients derived in (\ref{eq:wrt_thetai}), (\ref{eq:wrt_xi}), and (\ref{eq:thetag}), it is reasonable to apply the gradient descent-based algorithm to solve model (\ref{eq:3D-model}), despite the presence of nuclear norm terms. Specifically, the adaptive moment estimation (Adam) optimizer \cite{Cascarano2020CombiningWT} is employed to train our model.

\section{Experiments}
\label{sec:experiments}
To verify the effectiveness and performance improvement of the proposed method, experiments were conducted on simulated and real degraded HSI data, respectively, in this section. The pixel values of HSIs are normalized into [0, 1].
The related mainstream methods are employed for performance comparison with the proposed method: 
TNN \cite{Zhang2015ExactTC}  represents the conventional tensor nuclear norm method with the Fourier transform;
t-CTV\footnote{https://github.com/wanghailin97} \cite{Wang2023GuaranteedTR} fuses Low-rankness and Smoothness by a regularizer, tensor correlated total variation;
WSTNN\footnote{https://yubangzheng.github.io/} \cite{Zheng2018TensorNR} is the 3D form of the Fourier TNN;
S2NTNN\footnote{https://github.com/YisiLuo/S2NTNN} \cite{luo2022self} is a fully connected network transform-based TNN method;
CoNoT \cite{wang2022conot} is a coupled nonlinear transform-based TNN method that couples the spectral and the spatial transforms;
HIR-Diff\footnote{https://github.com/LiPang/HIRDiff} \cite{Pang2024HIRDiffUH} combines a diffusion technique with a low-rank tensor decomposition.
All deep learning-based models are trained on an NVIDIA L40 GPU. Other experiments are implemented on MATLAB (R2023b) using Windows 11 with an Intel Core i9-10900K 3.70GHz processor and 128-GB RAM.

\subsection{HSI Inpainting on Simulated Data}

\subsubsection{Datasets}
Three HSIs are chosen as our simulation datasets. The first HSI is selected from the Pavia Centre dataset with size of $200 \times$ $200 \times 80$. The second one is chosen from the Pavia University dataset\footnote{https://www.ehu.eus/ccwintco/index.php/Hyperspectral\_Remote\_Sensing\_\\Scenes}, with size of $256 \times 256 \times 103$. The third one is chosen from the Washington DC Mall dataset\footnote{http://lesun.weebly.com/hyperspectral-data-set.html}, with size of $256 \times 256 \times 191$.

\subsubsection{Missing Scenes}
We compared the proposed method with the baseline methods in the following four scenarios.

\textit{Case 1 (Point Missing):} Randomly select elements in an HSI tensor as the missing part. The missing rates are set as 0.85, 0.90, 0.95, and 0.99.

\textit{Case 2 (Stripe Slice Missing):} Randomly select vertical stripe
slice matrices in an HSI tensor as the missing part. The missing rates are set as 0.10, 0.30, and 0.50.

\begin{figure}[!htbp]
\centering
\subfigure[]{
\includegraphics[width=0.12\textwidth]{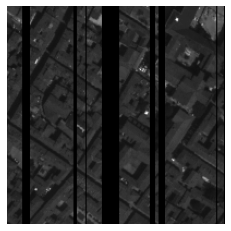}}\subfigure[]{\includegraphics[width=0.12\textwidth]{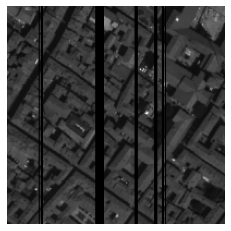}}\subfigure[]{\includegraphics[width=0.12\textwidth]{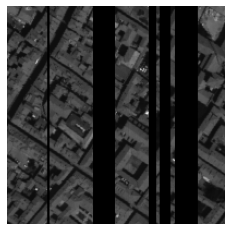}}\subfigure[]{\includegraphics[width=0.12\textwidth]{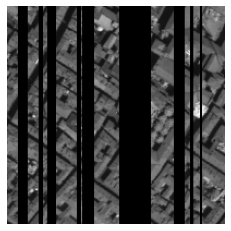}}
\caption{The four deadline patterns on the Pavia Centre dataset.
(a) The pattern on band 1 to 20, with its total width being 0.2 times the image width. (b) The pattern on band 21 to 40, with its total width being 0.1 times the image width. (c) The pattern on band 41 to 60, with its total width being 0.3 times the image width. (d) The pattern on band 61 to 80, with its total width being 0.4 times the image width.}
\label{fig:DLine}
\end{figure}

\textit{Case 3 (Deadline Missing on Consecutive Bands):} 4 deadline patterns are set on four groups of consecutive bands, respectively. Each band group has $\frac{1}{4}$ bands of the whole tensor. 
The missing rate (the total width of the deadlines/the image width) of 4 patterns are set as 0.1, 0.2, 0.3, 0.4, respectively. For each pattern, the number of deadlines is randomly selected from {6,7,...,10}. An example to illustrate this kind of missing is shown in Fig. \ref{fig:DLine}.

\textit{Case 4 (Case 1 + Case 2 + Case 3):} The missing part of an HSI tensor consists of 3 parts: Case 1 with the missing rate = 0.9, Case 2 with the missing rate = 0.1, and Case 3.

\subsubsection{Parameter settings}
In our model, the hyperparameters include dimensions in the latent spaces $\{\widetilde{n_i}\}_{i=1}^3$, weights on different directions $\{\alpha_i\}_{i=1}^3$, and regularization parameters $\{\lambda_i\}_{i=1}^3$. Specifically, we set $\widetilde{n_i} = n_i,\, i=1,2,3$, i.e., the latent space dimensions are the same as the image domain ones. For the direction weights, we set $\left(\alpha_1, \alpha_2, \alpha_3\right)=\beta \frac{(1,1, \theta)}{2+\theta}$. Then the three parameters $\{\alpha_i\}_{i=1}^3$  are converted into two: $\beta$ and $\theta$. For the regularization parameters $\{\lambda_i\}_{i=1}^3$, to reduce the computational cost of hyperparameter learning, we set
$\alpha_i\lambda_i=\gamma_i=\gamma$
, which yields $\lambda_i=\frac{\gamma}{\alpha_i}$, $i=1,\,2,\,3$. As a result, only a single hyperparameter $\gamma$ needs to be tuned. We adjust hyperparameters $\beta$, $\theta$, $\lambda$ to achieve the highest PSNR value,
with $\beta$ controlling the importance of the low-rankness regularizers, $\theta$ balancing the importance of low-rankness between the spatial mode (mode-1 and mode-2) and the spectral mode (mode-3), $\gamma$ determining the importance of fidelity of $\mathcal{X}_i$'s, as illustrated in (\ref{eq:hyperPara}): 
\begin{equation}
    \begin{aligned}
\mathcal{L} = & \ \beta \sum_{i = 1}^3 \widetilde{ \alpha_i} \sum_{k = 1}^{\widetilde{n_i}}\left\|\left(\widehat{\mathcal{X}}_i\right)_i^{(k)}\right\|_*
\nonumber\\
& \ +\gamma \sum_{i = 1}^3 \left\|\mathcal{P}_{\Omega}\left(\mathcal{X}_i-\mathcal{O}\right)\right\|_F^2
+\left\|\mathcal{P}_{\Omega}(\mathcal{X}-\mathcal{O})\right\|_F^2,
\end{aligned} \label{eq:hyperPara}
\end{equation}
where $\mathcal{L}$ is the objective function in (\ref{eq:3D-model}), $(\widetilde{\alpha_1},\widetilde{\alpha_2},\widetilde{\alpha_3})=\frac{(1,1,\theta)}{2+\theta}$.
As for the comparison methods, all relevant hyperparameters are manually adjusted following the authors' suggestions.

Considering the initialization of learnable parameters, for the neural network parameters $\{\Theta_i\}_{i=1}^3$ and $\Theta_G$, the default normal distribution in Pytorch\footnote{https://pytorch.org/docs/stable/nn.init.html} is utilized to initialize them. For the latent space tensors $\{\widehat{\mathcal{X}}_i\}_{i=1}^3$, since we assign the dimensions of latent space tensors the same as the corresponding images to be recovered, we initialize the $\{\widehat{\mathcal{X}}_i\}_{i=1}^3$ by the conventional TNN method \cite{Zhang2015ExactTC}.

\subsubsection{Quantitative Indices} We quantitatively evaluate the inpainting results of different methods using three common HSI restoration indices: peak signal-to-noise ratio (PSNR), structural similarity (SSIM), and spectral angle mapper (SAM). PSNR and SSIM are spatial information-based evaluation indices, while SAM is employed to evaluate spectral information. Higher PSNR and SSIM values and lower SAM values indicate better inpainting performance.
\subsubsection{Experimental Results on Pavia Centre Dataset}

\begin{figure*}[!htbp]
    \centering
    \includegraphics[width=1\textwidth]{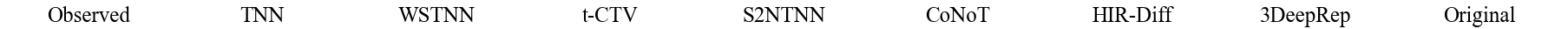}
    \includegraphics[width=1\textwidth]{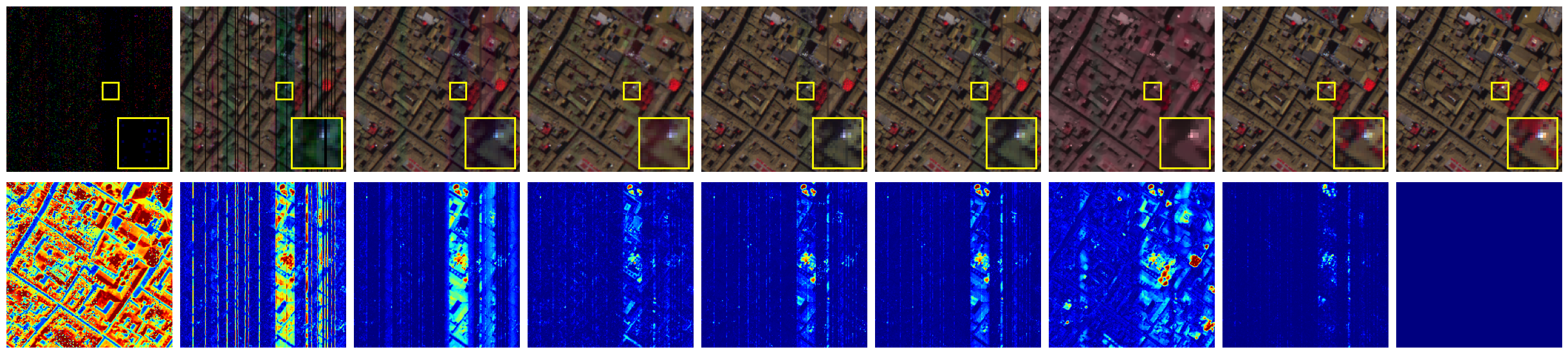}
    \includegraphics[width=0.8\textwidth]{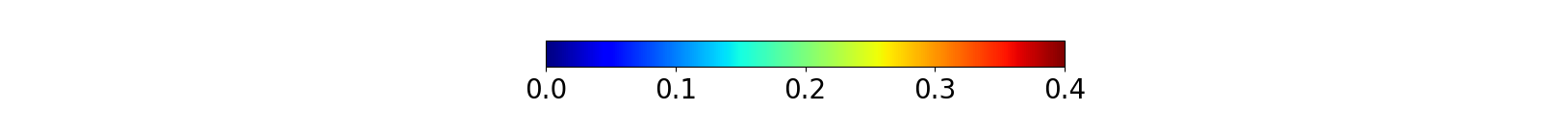}
    \caption{Inpainting results of the Pavia Centre dataset under Case 4 (mixture missing). The top row is the false-color images composed of bands (R: 70, G: 40, and B: 10). The bottom row is the absolute error maps between the ground truth and the recovered images (on band 58).}
\label{fig:Pavia}
\end{figure*}

\textit{Qualitative Comparison:}
Fig. \ref{fig:Pavia}  is the visualized inpainting performance comparison results for the  Pavia Centre dataset, where the upper row shows the recovered false-color images, while the lower row presents the absolute error maps. As observed from the false-color images, 3DeepRep exhibits the most similar color tone and the clearest details compared to the ground truth (e.g., the red dots within the zoomed-in yellow square). In the error maps, the result from 3DeepRep appears closest to pure dark blue, indicating the highest similarity between the reconstructed image and the ground truth. In contrast, the false-color images recovered by TNN, WSTNN, S2NTNN, and CoNoT exhibit noticeable stripe noise, indicating insufficient exploitation of local spatial information. The result of t-CTV shows blurred details, as evident in the zoomed-in region. Additionally, the overall color tone of the HIR-Diff result deviates from that of the ground truth, suggesting a significant spectral distortion.

\textit{Quantitative Comparison:}
Table \ref{tab:Pavia} lists the quantitative inpainting results of 3DeepRep for the Pavia Centre dataset, comparing with the 6 baseline methods under the 4 missing cases, with boldface marking the best results and underline marking the second-best ones. In general, 3DeepRep outperforms all six baseline methods across the four missing cases. The only exception occurs in Case 2, MR = 0.10, where the SAM value of 3DeepRep is inferior to that of CoNoT. However, in this scenario, 3DeepRep achieves a substantially higher PSNR and a superior SSIM compared to CoNoT, demonstrating its overall better inpainting performance. It is worth noting that in Case 3 (wide deadline missing), the PSNR of 3DeepRep exceeds that of the second-best method, CoNoT, by 7.553 dB. This is a relatively large margin, highlighting the superiority of the 3D structure of the proposed method.

\begin{table*}[!htbp]
\centering
\caption{Quantitative Comparison of Different Comparison Methods Under Different Inpainting Cases on Pavia Centre Dataset. \textbf{Boldface} and \underline{Underline} Highlight the Best and Second-Best Values, Respectively.}
\label{tab:Pavia}
\begin{tabular}{ccccccccccc}
\toprule[1.5pt]
Case                     & MR                    & Metric & Observed & TNN    & WSTNN       & t-CTV        & S2NTNN      & CoNoT          & HIR-Diff & 3DeepRep            \\ \hline
\multirow{12}{*}{Case 1} & \multirow{3}{*}{0.99} & PSNR   & 12.013   & 12.013 & 17.552      & 26.352       & 26.001      & {\ul 29.523}   & 18.870   & \textbf{30.754} \\
                         &                       & SSIM   & 0.004    & 0.004  & 0.243       & 0.752        & 0.755       & {\ul 0.885}    & 0.513    & \textbf{0.915}  \\
                         &                       & SAM    & 2.207    & 2.207  & 0.178       & 0.128        & 0.090       & {\ul 0.075}    & 0.452    & \textbf{0.056}  \\ \cline{2-11} 
                         & \multirow{3}{*}{0.95} & PSNR   & 12.192   & 28.879 & 37.296      & 33.786       & 37.932      & {\ul 40.456}   & 26.821   & \textbf{43.107} \\
                         &                       & SSIM   & 0.018    & 0.860  & 0.981       & 0.950        & 0.981       & {\ul 0.989}    & 0.867    & \textbf{0.994}  \\
                         &                       & SAM    & 1.381    & 0.126  & 0.034       & 0.074        & 0.033       & {\ul 0.026}    & 0.162    & \textbf{0.019}  \\ \cline{2-11} 
                         & \multirow{3}{*}{0.90} & PSNR   & 12.426   & 32.637 & 45.952      & 38.717       & 46.416      & {\ul 48.980}   & 28.128   & \textbf{51.547} \\
                         &                       & SSIM   & 0.035    & 0.930  & 0.997       & 0.981        & 0.997       & {\ul 0.998}    & 0.887    & \textbf{0.999}  \\
                         &                       & SAM    & 1.255    & 0.097  & 0.013       & 0.047        & 0.014       & {\ul 0.012}    & 0.128    & \textbf{0.009}  \\ \cline{2-11} 
                         & \multirow{3}{*}{0.85} & PSNR   & 12.676   & 35.520 & 52.911      & 42.773       & 52.127      & {\ul 54.990}   & 28.480   & \textbf{58.660} \\
                         &                       & SSIM   & 0.053    & 0.957  & {\ul 0.999} & 0.991        & {\ul 0.999} & {\ul 0.999}    & 0.891    & \textbf{1.000}  \\
                         &                       & SAM    & 1.177    & 0.078  & 0.007       & 0.033        & 0.009       & {\ul 0.006}    & 0.119    & \textbf{0.003}  \\ \hline
\multirow{9}{*}{Case 2}  & \multirow{3}{*}{0.50} & PSNR   & 14.940   & 14.940 & 26.319      & {\ul 28.560} & 23.178      & 25.963         & 28.428   & \textbf{30.201} \\
                         &                       & SSIM   & 0.216    & 0.216  & 0.772       & {\ul 0.857}  & 0.588       & 0.766          & 0.835    & \textbf{0.904}  \\
                         &                       & SAM    & 1.571    & 1.157  & {\ul 0.042} & 0.047        & 0.067       & 0.069          & 0.075    & \textbf{0.036}  \\ \cline{2-11} 
                         & \multirow{3}{*}{0.30} & PSNR   & 17.164   & 17.164 & 30.642      & {\ul 32.187} & 29.789      & 31.216         & 31.031   & \textbf{36.912} \\
                         &                       & SSIM   & 0.381    & 0.381  & 0.913       & {\ul 0.939}  & 0.897       & 0.926          & 0.908    & \textbf{0.977}  \\
                         &                       & SAM    & 0.942    & 0.607  & 0.022       & 0.025        & 0.032       & {\ul 0.021}    & 0.044    & \textbf{0.019}  \\ \cline{2-11} 
                         & \multirow{3}{*}{0.10} & PSNR   & 22.137   & 22.137 & 39.751      & 39.270       & 39.610      & {\ul 43.017}   & 34.780   & \textbf{47.956} \\
                         &                       & SSIM   & 0.703    & 0.703  & 0.989       & 0.988        & 0.989       & {\ul 0.995}    & 0.956    & \textbf{0.998}  \\
                         &                       & SAM    & 0.298    & 0.235  & {\ul 0.005}       & 0.007        & 0.012       & \textbf{0.004} & 0.038    & 0.011           \\ \hline
\multicolumn{2}{c}{\multirow{3}{*}{Case 3}}      & PSNR   & 18.415   & 27.614 & 39.722      & 35.799       & 39.754      & {\ul 43.499}   & 29.682   & \textbf{51.052} \\
\multicolumn{2}{c}{}                             & SSIM   & 0.648    & 0.889  & 0.988       & 0.959        & 0.990       & {\ul 0.995}    & 0.901    & \textbf{0.999}  \\
\multicolumn{2}{c}{}                             & SAM    & 0.470    & 0.139  & 0.024       & 0.064        & 0.016       & {\ul 0.012}    & 0.107    & \textbf{0.007}  \\ \hline
\multicolumn{2}{c}{\multirow{3}{*}{Case 4}}      & PSNR   & 12.274   & 20.251 & 30.421      & 31.890       & 32.644      & {\ul 33.279}   & 26.880   & \textbf{38.610} \\
\multicolumn{2}{c}{}                             & SSIM   & 0.024    & 0.583  & 0.934       & 0.930        & 0.951       & {\ul 0.958}    & 0.851    & \textbf{0.986}  \\
\multicolumn{2}{c}{}                             & SAM    & 1.512    & 0.333  & 0.085       & 0.078        & {\ul 0.042} & {\ul 0.042}    & 0.147    & \textbf{0.022} \\         
\bottomrule[1.5pt]
\end{tabular}
\end{table*}

\subsubsection{Experimental Results on Pavia University Dataset}

\begin{figure*}[!htbp]
    \centering
    \includegraphics[width=1\textwidth]{pics/titles.png}
    \includegraphics[width=1\textwidth]{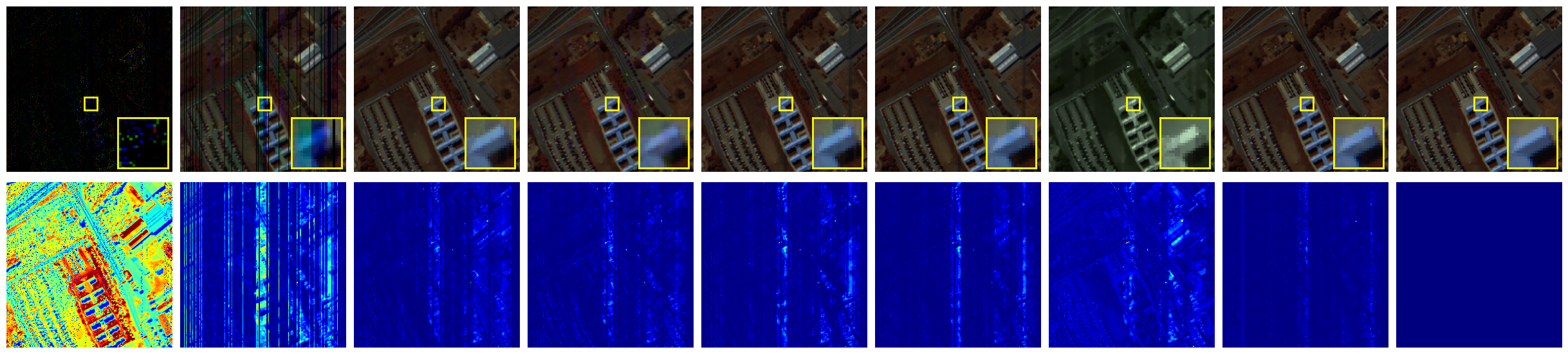}
    \includegraphics[width=0.8\textwidth]{pics/color_bar.png}
    \caption{Inpainting results of the Pavia University dataset under Case 4 (mixture missing). The top row is the false-color images composed of bands (R: 70, G: 40, and B: 10). The bottom row is the absolute error maps between the ground truth and the recovered images (on band 75).}
\label{fig:PaviaU}
\end{figure*}

\textit{Qualitative Comparison:}
Fig. \ref{fig:PaviaU} is the visual result of the inpainting model comparison on the Pavia University dataset. As shown in Fig. \ref{fig:PaviaU},  the proposed 3DeepRep achieves the best overall inpainting performance. It produces a color tone that closely aligns with the ground truth, preserves clear boundaries (observed in the zoomed-in region), and eliminates stripe and deadline noise. In contrast, TNN, S2NTNN, and CoNoT suffer from stripe or deadline noise. The result of t-CTV exhibits blurred edges, particularly visible in the enlarged area, while HIR-Diff shows a significant deviation in color tone.

\textit{Quantitative comparison:}
Table \ref{tab:PaviaU} is the quantitative model comparison result for the Pavia University dataset. In general, 3DeepRep outperforms all the baseline methods across all the scenarios, except that in a few scenes, its SSIM or SAM is slightly inferior to the best-performing method, while its PSNR remains the highest throughout.

\begin{table*}[!htbp]
\centering
\caption{Quantitative Comparison of Different Comparison Methods Under Different Inpainting Cases on Pavia University Dataset. \textbf{Boldface} and \underline{Underline} Highlight the Best and Second-Best Values, Respectively.}
\label{tab:PaviaU}
\begin{tabular}{ccccccccccc}
\toprule[1.5pt]
Case                     & MR                    & Metric & Observed & TNN    & WSTNN          & t-CTV        & S2NTNN       & CoNoT          & HIR-Diff & 3DeepRep            \\ \hline
\multirow{12}{*}{Case 1} & \multirow{3}{*}{0.99} & PSNR   & 14.010   & 14.010 & 18.738         & 27.384       & 27.220       & {\ul 30.774}   & 18.938   & \textbf{31.058} \\
                         &                       & SSIM   & 0.012    & 0.012  & 0.418          & 0.739        & 0.745        & \textbf{0.877} & 0.433    & {\ul 0.868}     \\
                         &                       & SAM    & 2.056    & 2.056  & 0.271          & 0.150        & 0.122        & {\ul 0.093}    & 0.460    & \textbf{0.092}  \\ \cline{2-11} 
                         & \multirow{3}{*}{0.95} & PSNR   & 14.189   & 28.565 & 36.651         & 33.778       & 36.697       & {\ul 38.725}   & 22.721   & \textbf{40.460} \\
                         &                       & SSIM   & 0.035    & 0.766  & 0.963          & 0.920        & 0.956        & {\ul 0.968}    & 0.643    & \textbf{0.971}  \\
                         &                       & SAM    & 1.364    & 0.162  & 0.054          & 0.090        & 0.058        & {\ul 0.052}    & 0.281    & \textbf{0.049}  \\ \cline{2-11} 
                         & \multirow{3}{*}{0.90}  & PSNR   & 14.425   & 32.629 & 42.285         & 37.965       & 42.194       & {\ul 43.167}   & 29.193   & \textbf{44.101} \\
                         &                       & SSIM   & 0.059    & 0.883  & \textbf{0.983} & 0.960        & 0.977        & {\ul 0.981}    & 0.850    & 0.980           \\
                         &                       & SAM    & 1.255    & 0.116  & \textbf{0.038} & 0.064        & 0.045        & {\ul 0.041}    & 0.138    & 0.042           \\ \cline{2-11} 
                         & \multirow{3}{*}{0.85} & PSNR   & 14.674   & 35.376 & {\ul 44.933}   & 40.892       & 44.654       & 44.887         & 29.869   & \textbf{45.441} \\
                         &                       & SSIM   & 0.081    & 0.925  & \textbf{0.987} & 0.974        & 0.982        & {\ul 0.985}          & 0.863    & 0.983           \\
                         &                       & SAM    & 1.177    & 0.093  & \textbf{0.032} & 0.051        & 0.039        & {\ul 0.036}          & 0.126    & 0.039           \\ \hline
\multirow{9}{*}{Case 2}  & \multirow{3}{*}{0.50}  & PSNR   & 16.969   & 16.969 & 27.869         & {\ul 30.472} & 24.803       & 27.907         & 30.452   & \textbf{33.080} \\
                         &                       & SSIM   & 0.252    & 0.252  & 0.819          & {\ul 0.877}  & 0.681        & 0.819          & 0.860    & \textbf{0.925}  \\
                         &                       & SAM    & 1.571    & 1.155  & 0.072          & {\ul 0.062}  & 0.090        & 0.066          & 0.084    & \textbf{0.055}  \\ \cline{2-11} 
                         & \multirow{3}{*}{0.30}  & PSNR   & 19.154   & 19.154 & 32.520         & {\ul 34.074} & 31.647       & 33.515         & 32.607   & \textbf{39.179} \\
                         &                       & SSIM   & 0.393    & 0.393  & 0.932          & {\ul 0.945}  & 0.910        & 0.942          & 0.903    & \textbf{0.975}  \\
                         &                       & SAM    & 0.933    & 0.608  & 0.032          & {\ul 0.031}  & 0.044        & \textbf{0.027} & 0.075    & 0.035           \\ \cline{2-11} 
                         & \multirow{3}{*}{0.10}  & PSNR   & 19.154   & 24.339 & 41.314         & 40.480       & 40.525       & {\ul 42.402}   & 34.962   & \textbf{44.789} \\
                         &                       & SSIM   & 0.393    & 0.707  & 0.990          & 0.987        & 0.985        & {\ul 0.991}    & 0.928    & \textbf{0.993}  \\
                         &                       & SAM    & 0.933    & 0.172  & {\ul 0.008}    & 0.009        & 0.024        & \textbf{0.007} & 0.070    & 0.017           \\ \hline
\multicolumn{2}{c}{\multirow{3}{*}{Case 3}}      & PSNR   & 20.779   & 25.773 & 37.672         & 33.391       & {\ul 39.818} & 39.790         & 29.860   & \textbf{42.889} \\
\multicolumn{2}{c}{}                             & SSIM   & 0.679    & 0.831  & 0.971          & 0.940        & 0.978        & {\ul 0.979}    & 0.866    & \textbf{0.990}  \\
\multicolumn{2}{c}{}                             & SAM    & 0.411    & 0.192  & 0.039          & 0.066        & 0.031        & {\ul 0.028}    & 0.130    & \textbf{0.023}  \\ \hline
\multicolumn{2}{c}{\multirow{3}{*}{Case 4}}      & PSNR   & 14.277   & 21.620 & 33.355         & 30.665       & 32.999       & {\ul 34.434}   & 26.536   & \textbf{38.057} \\
\multicolumn{2}{c}{}                             & SSIM   & 0.043    & 0.542  & 0.921          & 0.875        & 0.914        & {\ul 0.934}    & 0.810    & \textbf{0.959}  \\
\multicolumn{2}{c}{}                             & SAM    & 1.494    & 0.402  & 0.075          & 0.122        & 0.079        & {\ul 0.074}    & 0.208    & \textbf{0.054} \\ \bottomrule[1.5pt]
\end{tabular}
\end{table*}

\subsubsection{Experimental Results on Washington DC Mall Dataset}

\begin{figure*}[!htbp]
    \centering
    \includegraphics[width=1\textwidth]{pics/titles.png}
    \includegraphics[width=1\textwidth]{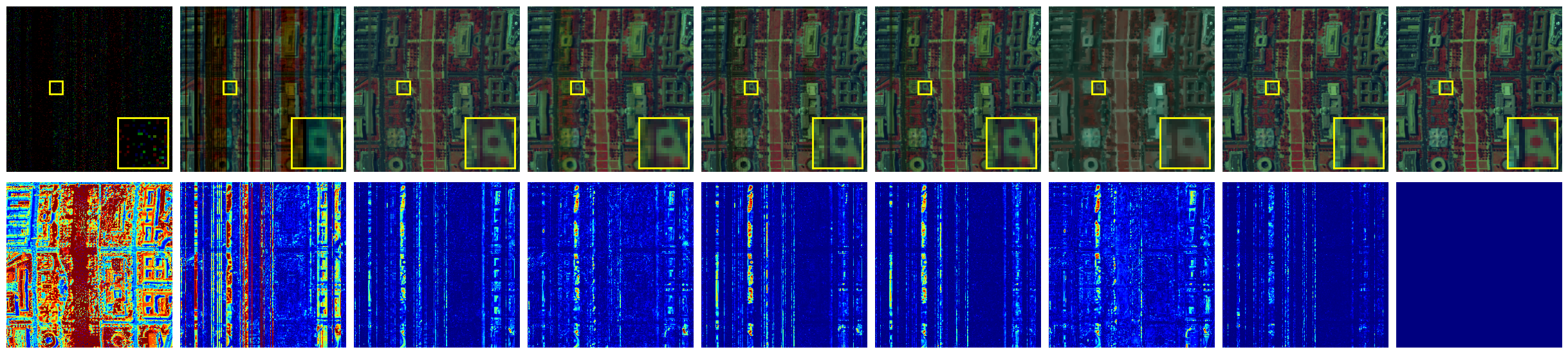}
    \includegraphics[width=0.8\textwidth]{pics/color_bar.png}
    \caption{Inpainting results of the Washington DC Mall dataset under Case 4 (mixture missing). The top row is the false-color images composed of bands (R: 70, G: 40, and B: 10). The bottom row is the absolute error maps between the ground truth and the recovered images (on band 60).}
\label{fig:dc}
\end{figure*}

\textit{Qualitative Comparison:}
Fig. \ref{fig:dc} presents the inpainting results of the comparison methods on the Washington DC Mall dataset under Case 4. As observed, 3DeepRep exhibits the sharpest details and produces an error map most closely aligned with the ground truth one. In contrast, TNN suffers from prominent stripe noise, while WSTNN, t-CTV, S2NTNN, and CoNoT display less distinct detail recovery—for example, the red circles in the zoomed-in regions of these methods are less vivid than those in the 3DeepRep result. HIR-Diff still exhibits a noticeable color tone deviation. It is worth noting that compared to the visual results on the Pavia Centre and Pavia University datasets, WSTNN, S2NTNN, and CoNoT display weaker stripe and deadline noise on the Washington DC Mall dataset. This improvement can be attributed to the higher number of spectral bands in the Washington DC Mall dataset (Pavia Centre: 80 bands; Pavia University: 103 bands; Washington DC Mall: 191 bands), which provides more information to support image reconstruction.

\textit{Quantitative Comparison:}
Table \ref{tab:dc} lists the quantitative results of model comparison for the Washington DC Mall dataset. As observed, overall, 3DeepRep outperforms all the baseline methods, except that in Case 2, MR = 0.50, its PSNR and SSIM are slightly lower than those of the top-performing method HIR-Diff, and its SAM is slightly higher than that of t-CTV. In all other scenes, 3DeepRep achieves the highest PSNR values.
\begin{table*}[!htbp]
\centering
\caption{Quantitative Comparison of Different Comparison Methods Under Different Inpainting Cases on Washington DC Mall. \textbf{Boldface} and \underline{Underline} Highlight the Best and Second-Best Values, Respectively.}
\label{tab:dc}
\begin{tabular}{ccccccccccc}
\toprule[1.5pt]
Case                     & MR                    & Metric & Observed & TNN    & WSTNN          & t-CTV          & S2NTNN      & CoNoT          & HIR-Diff        & 3DeepRep            \\ \hline
\multirow{12}{*}{Case 1} & \multirow{3}{*}{0.99} & PSNR   & 25.615   & 25.615 & 31.825         & 33.404         & 40.718      & {\ul 43.150}   & 31.019          & \textbf{43.282} \\
                         &                       & SSIM   & 0.266    & 0.266  & 0.604          & 0.776          & 0.908       & {\ul 0.940}    & 0.660           & \textbf{0.949}  \\
                         &                       & SAM    & 1.725    & 1.725  & 0.295          & 0.175          & 0.105       & {\ul 0.090}    & 0.515           & \textbf{0.067}  \\ \cline{2-11} 
                         & \multirow{3}{*}{0.95} & PSNR   & 25.794   & 33.314 & 48.903         & 38.670         & 50.505      & {\ul 53.151}   & 35.608          & \textbf{54.433} \\
                         &                       & SSIM   & 0.289    & 0.811  & 0.986          & 0.934          & 0.990       & {\ul 0.994}    & 0.836           & \textbf{0.995}  \\
                         &                       & SAM    & 1.355    & 0.155  & 0.046          & 0.090          & 0.032       & {\ul 0.029}    & 0.277           & \textbf{0.025}  \\ \cline{2-11} 
                         & \multirow{3}{*}{0.90}  & PSNR   & 25.615   & 36.826 & {\ul 56.240}   & 42.567         & 40.325      & 43.080         & 36.681          & \textbf{59.309} \\
                         &                       & SSIM   & 0.266    & 0.902  & {\ul 0.998}    & 0.972          & 0.911       & 0.939          & 0.858           & \textbf{0.999}  \\
                         &                       & SAM    & 1.725    & 0.106  & {\ul 0.020}    & 0.058          & 0.097       & 0.090          & 0.242           & \textbf{0.015}  \\ \cline{2-11} 
                         & \multirow{3}{*}{0.85} & PSNR   & 26.276   & 39.481 & {\ul 60.024}   & 45.403         & 57.920      & 59.377         & 37.779          & \textbf{61.130} \\
                         &                       & SSIM   & 0.340    & 0.941  & \textbf{0.999} & 0.984          & 0.998       & \textbf{0.999} & 0.888           & \textbf{0.999}  \\
                         &                       & SAM    & 1.177    & 0.081  & \textbf{0.013} & 0.043          & 0.014       & 0.017          & 0.196           & \textbf{0.013}  \\ \hline
\multirow{9}{*}{Case 2}  & \multirow{3}{*}{0.50}  & PSNR   & 28.629   & 28.629 & 37.956         & 36.657         & 35.057      & 35.650         & \textbf{38.755} & {\ul 38.696}    \\
                         &                       & SSIM   & 0.512    & 0.512  & 0.855          & {\ul 0.883}    & 0.762       & 0.793          & \textbf{0.888}  & 0.880           \\
                         &                       & SAM    & 1.571    & 1.148  & 0.089          & \textbf{0.068} & 0.138       & 0.089          & 0.091           & {\ul 0.083}     \\ \cline{2-11} 
                         & \multirow{3}{*}{0.30}  & PSNR   & 30.676   & 30.676 & {\ul 41.415}   & 40.448         & 39.123      & 40.559         & 40.391          & \textbf{41.855} \\
                         &                       & SSIM   & 0.634    & 0.634  & 0.931          & \textbf{0.946} & 0.900       & 0.920          & 0.927           & {\ul 0.941}     \\
                         &                       & SAM    & 0.933    & 0.601  & 0.045          & \textbf{0.034} & 0.063       & {\ul 0.040}    & 0.075           & {\ul 0.040}     \\ \cline{2-11} 
                         & \multirow{3}{*}{0.10}  & PSNR   & 35.198   & 35.198 & 47.103         & 46.431         & 44.160      & {\ul 47.379}   & 42.199          & \textbf{49.459} \\
                         &                       & SSIM   & 0.850    & 0.850  & 0.983          & {\ul 0.987}    & 0.974       & 0.984          & 0.956           & \textbf{0.991}  \\
                         &                       & SAM    & 0.307    & 0.215  & 0.012          & \textbf{0.010} & 0.026       & {\ul 0.011}    & 0.064           & 0.017           \\ \hline
\multicolumn{2}{c}{\multirow{3}{*}{Case 3}}      & PSNR   & 31.983   & 29.252 & {\ul 47.561}   & 36.035         & 46.106      & 46.120         & 38.174          & \textbf{47.977} \\
\multicolumn{2}{c}{}                             & SSIM   & 0.748    & 0.773  & 0.962          & 0.913          & {\ul 0.967} & 0.959          & 0.879           & \textbf{0.980}  \\
\multicolumn{2}{c}{}                             & SAM    & 0.514    & 0.407  & {\ul 0.047}    & 0.103          & 0.054       & 0.057          & 0.144           & \textbf{0.033}  \\ \hline
\multicolumn{2}{c}{\multirow{3}{*}{Case 4}}      & PSNR   & 25.878   & 27.151 & {\ul 42.863}   & 34.989         & 41.147      & 42.730         & 36.309          & \textbf{46.337} \\
\multicolumn{2}{c}{}                             & SSIM   & 0.297    & 0.621  & 0.944          & 0.884          & 0.937       & {\ul 0.949}    & 0.857           & \textbf{0.977}  \\
\multicolumn{2}{c}{}                             & SAM    & 1.497    & 0.504  & {\ul 0.070}    & 0.121          & 0.092       & 0.078          & 0.199           & \textbf{0.036}  \\ \bottomrule[1.5pt]
\end{tabular}
\end{table*}

\subsection{HSI Inpainting on Real Data} 
\begin{figure*}[!htbp]
    \centering
    \includegraphics[width=1\textwidth]{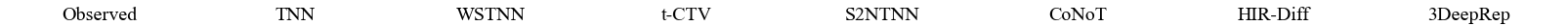}
    \includegraphics[width=1\textwidth]{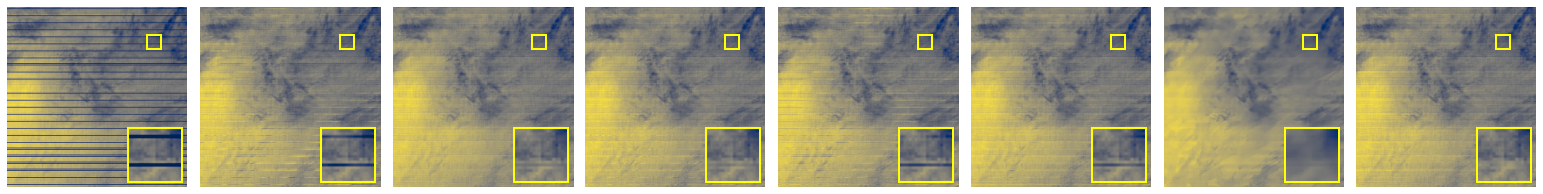}
    \includegraphics[width=0.3\textwidth]{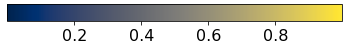}
    \caption{The inpainting results on Band 27 of the chosen MODIS Level-1B HSI dataset which is with the 1 km resolution at Antarctic Peninsula on February 26, 2011.}
\label{fig:real}
\end{figure*}

In this section, we adopt a MODIS Level-1B dataset with the 1 km resolution\footnote{https://ladsweb.modaps.eosdis.nasa.gov/missions-and-measurements/products/MOD021KM} as the testing dataset. The specific HSI chosen is the one at Antarctic Peninsula on February 26, 2011 consisting of emissive bands with size of $256 \times 256 \times 10$. The missing situation is that only Band 27 is found with missing pixel values out of the 10 chosen bands. And the missing pattern on Band 27 is illustrated in the leftmost picture in Fig. \ref{fig:real} which is a kind of strip missing.

The proposed 3DeepRep method as well as other comparison methods were conducted on the MODIS dataset. The hyperparameters of all the above methods were adjusted to achieve the best visual result. The inpainting results is illustrated in Fig. \ref{fig:real}. Notably,  only Band 27 is shown in Fig. \ref{fig:real} since it is the only corrupted band in the dataset although the above methods were conducted on the whole $256 \times 256 \times 10$ dataset. From Fig. \ref{fig:real}, it can be observed from the enlarged yellow square that 3DeepRep deletes the horizontal strips most clearly while TNN, S2NTNN, CoNoT that are all 1D methods retain the strips obviously. What' more, WSTNN and t-CTV still left a slight strip at the lower middle of the yellow square and HIR-Diff results in an overly smoothed and blurry image. In summary, the visual results demonstrated the superiority of 3DeepRep over the comparison methodds.

\subsection{Discussion}
\textit{1) Effectiveness of the Three-directional Transformation:} 
To evaluate the effectiveness of the proposed 3-directional transform, we compare it against three types of 1D configurations, where the deep transform is applied along a single direction, including mode-1, mode-2, or mode-3, by setting $i=1,2,3$ in the model formulation (\ref{eq:modei-model}), respectively. The comparison results are reported in the first block of Table~\ref{tab:discussion}. As observed, the 3D transform significantly boosts the reconstruction performance compared to its 1D counterparts. Interestingly, among the three 1D variants, the spectral transform (mode-3) outperforms the spatial transforms (mode-1 and mode-2), suggesting its stronger capacity in capturing informative structures in hyperspectral data.

\textit{2) Learnable Latent Tensors \textbf{vs.} Generated Latent Tensors:} 
In prior deep transform-based TNN methods \cite{luo2022self,wang2022conot}, the low-rank latent tensors are generated by neural networks. In contrast, our proposed 3DeepRep model treats the latent tensors $\mathcal{\widehat{X}}_i$ as learnable variables directly. To evaluate the effectiveness of this design, we compare it with a variant where each latent tensor is generated by a network, i.e., $\mathcal{\widehat{X}}_i$ is replaced by $f_i(\mathcal{X}_i^0)$ in the formulation of Eq.~(\ref{eq:3D-model}). The corresponding optimization problem is given by:
\begin{equation*}
    \min_{\Theta_i^{f,g},\, \mathcal{X}_i^0,\, \Theta_G} \sum_{i=1}^3 \alpha_i \mathcal{L}_i(\Theta_i, \mathcal{X}_i^0)
    + \left\| \mathcal{P}_{\Omega}\left(G(\mathcal{X}_1, \mathcal{X}_2, \mathcal{X}_3) - \mathcal{O} \right) \right\|_F^2,
\end{equation*}
where $\{\mathcal{X}_i^0\}_{i=1}^3 \subset \mathbb{R}^{n_1 \times n_2 \times n_3}$ are the inputs to the latent tensor generators, and $f_i: \mathbb{R}^{n_1 \times n_2 \times n_3} \rightarrow \mathbb{R}^{n_1 \times n_2 \times n_3}$ denotes the generator for the $i$-th latent tensor. Each generator $f_i(\cdot)$ shares the same network architecture as the corresponding transform $g_i(\cdot)$, and the final output tensors are given by $\mathcal{X}_i = g_i(f_i(\mathcal{X}_i^0))$. Here, $\Theta_i^{f,g}$ denote the learnable parameters in $f_i(\cdot)$ and $g_i(\cdot)$, for $i = 1, 2, 3$.

The comparison results are presented in the second block of Table~\ref{tab:discussion}. As observed, the learnable-latent-tensor design not only results in fewer parameters but also leads to significantly better inpainting performance compared to the generated-latent-tensor representation methods.

\begin{table}[!htbp]
\centering
\caption{Quantitative Results of Different Inpainting Methods under Case 4 on the Pavia Centre Dataset. The Best Value Is Marked by Boldface.}
\label{tab:discussion}
\setlength{\tabcolsep}{2pt} 
\begin{tabular}{cccccc}
\toprule[1.5pt]
                                                                                                 & Method               & PSNR                 & SSIM                 & SAM                  & Parameter size        \\ \hline
\multirow{4}{*}{Direction}                & Mode-1               & 30.838               & 0.925                & 0.077                & 3.242M              \\
                                                                                                 & Mode-2               & 21.907               & 0.647                & 0.266                & 3.242M              \\
                                                                                                 & Mode-3               & 33.939               & 0.961                & 0.036                & \textbf{3.207M}     \\
                                                                                                 & 3D                   & \textbf{38.610}      & \textbf{0.986}       & \textbf{0.022}       & 9.712M              \\ \hline                                                                                                 
\multirow{2}{*}{\begin{tabular}[c]{@{}c@{}}Acquisition of\\ latent tensors\end{tabular}}          & Generated          & 36.823               & 0.980                & 0.025                & 9.803M              \\
                                                                                                 & Learnable         & \textbf{38.610}      & \textbf{0.986}       & \textbf{0.022}       & \textbf{9.712M}     \\ \hline

\multirow{4}{*}{\begin{tabular}[c]{@{}c@{}}Expansion Ratio \\ of the Latent Spaces\end{tabular}} & $k$=1                  & 38.610               & \textbf{0.986}       & \textbf{0.022}       & \textbf{9.712M}     \\
                                                                                                 & $k$=2                  & \textbf{38.913}      & \textbf{0.986}       & \textbf{0.022}       & 19.403M             \\
                                                                                                 & $k$=3                  & 38.601               & \textbf{0.986}       & \textbf{0.022}       & 29.094M             \\
                                                                                                 & $k$=4                  & 38.310               & 0.984                & \textbf{0.022}       & 38.784M             \\  \bottomrule[1.5pt]
\multicolumn{1}{l}{}                                                                             & \multicolumn{1}{l}{} & \multicolumn{1}{l}{} & \multicolumn{1}{l}{} & \multicolumn{1}{l}{} & \multicolumn{1}{l}{} \\
\multicolumn{1}{l}{}                                                                             & \multicolumn{1}{l}{} & \multicolumn{1}{l}{} & \multicolumn{1}{l}{} & \multicolumn{1}{l}{} & \multicolumn{1}{l}{}
\end{tabular}
\end{table}

\textit{3) Hyperparameter Analysis:} 
We analyze the sensitivity of the hyperparameters involved in the proposed method on Case 4 (mixture missing) across all three datasets. The hyperparameters to analysize include dimensions in the latent spaces $\{\widetilde{n_i}\}_{i=1}^3$, parameters to adjust weights for the three directions, $\beta$ and $\theta$, as well as $\gamma$ which controls the importance of the data fidelity terms of intermediate layers.

For $\{\widetilde{n_i}\}_{n=1}^3$, we set $\widetilde{n_i} = k n_i$, where $k \ge 1$ is used to provide redundancy \cite{Jiang2020DictionaryLW}. As shown in Table \ref{tab:discussion}, the inpainting performance—measured by PSNR, SSIM, and SAM—remains relatively stable across values of $k$ from 1 to 4. Therefore, to reduce computational cost, we adopt $k=1$ for experiments. It is worth noting that the highest PSNR is achieved at $k=2$, indicating that further parameter expansion does not necessarily enhance restoration performance.

\begin{figure}[!htbp]
    \centering
    \includegraphics[width=0.16\textwidth]{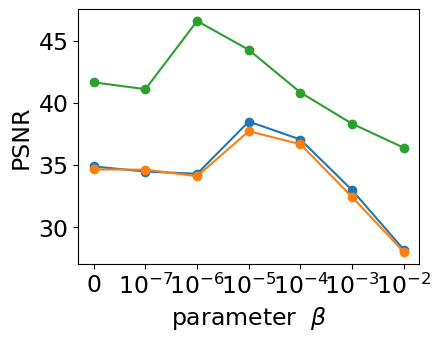}\includegraphics[width=0.16\textwidth]{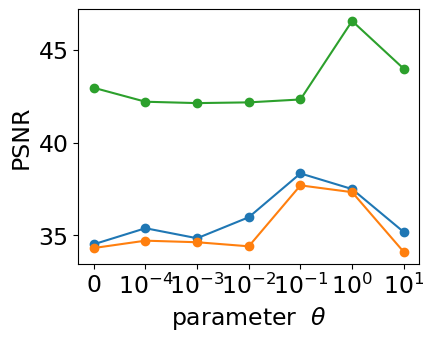}\includegraphics[width=0.16\textwidth]{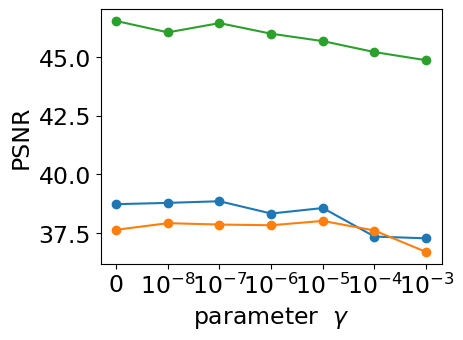}
    \includegraphics[width=0.49\textwidth]{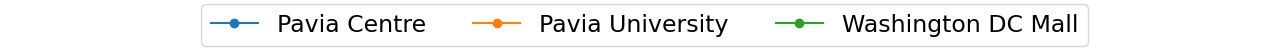}
    \caption{Sensitivity analysis of hyperparameters.}
\label{fig:hyp}
\end{figure}

The hyperparameter $\beta$ controls the importance of the low-rankness regularizers through the euqation $\left(\alpha_1, \alpha_2, \alpha_3\right)=\beta \frac{(1,1, \theta)}{2+\theta}$. As illustrated in Fig. \ref{fig:hyp}, 3DeepRep's inpainting performance is sensitive to $\beta$, indicating the importance of the low-rankness regularizers. The three datasets—Pavia Centre, Pavia University, and Washington DC Mall—reach their highest PSNRs at $\beta = 10^{-5}, 10^{-5}$, and $10^{-6}$, respectively.

 The hyperparameter $\theta$ balances the low-rankness importance between the spatial modes (mode-1 and mode-2) and the spectral mode (mode-3). The sensitivity analysis of $\theta$ is presented in Fig. \ref{fig:hyp}. As observed, the model performance is also sensitive to $\theta$. The three datasets, Pavia Centre, Pavia University, and Washington DC Mall, achieve their highest PSNRs at $\theta = 10^{-1}, 10^{-1}$, and $10^{0}$, respectively.

The hyperparameter $\gamma$ determines the importance of the data fidelity of $\{\mathcal{X}_i\}_{i=1}^3$. It is found in Fig. \ref{fig:hyp} that the model performance is not much sensitive to $\gamma$, indicating the data fidelity property is relatively less critical in the intermediate layers. The optimal values of $\gamma$ for the Pavia Centre dataset and the Pavia University dataset are $\gamma = 10^{-7}$ and $10^{-8}$, respectively. As for the Washington DC Mall, the optimal $\gamma$ is even zero.
\section{Conclusion} \label{sec:conclusion}
To address the HSI inpainting problem, this paper proposes a \textbf{3}-directional \textbf{deep} low-rank tensor \textbf{rep}resentation (3DeepRep) method that simultaneously extracts low-rank structures and characterizes cross-mode correlations along all three tensor modes. By leveraging a multi-directional deep transform framework, the proposed model fully exploits the spatial and spectral dependencies inherent in hyperspectral data, resulting in significantly improved inpainting accuracy. An efficient gradient-based optimization algorithm is developed to solve the model in a self-supervised manner, tailored to the structure of neural networks. 
The effectiveness and robustness of the proposed method are validated through experiments on both simulated and real HSI data, where it consistently outperforms existing approaches.
For future work, considering the high computational cost associated with nuclear norm minimization, low-rank tensor decomposition techniques may be explored as a more efficient alternative for enforcing low-rankness.
\section{Acknowledgment}
The authors would like to thank Xi-le Zhao for his valuable guidance and insightful suggestions throughout the development of this work. The authors also appreciate the contributions of researchers who generously shared their code, which facilitated comparison and reproducibility in our experimental evaluations.

\bibliographystyle{ieeetr}
\bibliography{ref}
\end{document}